\documentclass{article} 
\usepackage{iclr2027_conference,times}

\usepackage{amsmath,amsfonts,bm}

\def\figref#1{Figure~\ref{#1}}

\def\1{\bm{1}}

\DeclareMathAlphabet{\mathsfit}{\encodingdefault}{\sfdefault}{m}{sl}
\SetMathAlphabet{\mathsfit}{bold}{\encodingdefault}{\sfdefault}{bx}{n}

\def\gG{{\mathcal{G}}}

\newcommand{\R}{\mathbb{R}}

\DeclareMathOperator*{\argmin}{arg\,min}

\DeclareMathOperator{\sign}{sign}

\usepackage{graphicx}
\usepackage{subcaption}
\usepackage{booktabs}
\usepackage{nicefrac}
\usepackage{dsfont}
\usepackage{siunitx}
\DeclareSIUnit{\token}{token}
\usepackage{xcolor}
\usepackage{colortbl}
\usepackage{tabularx}
\usepackage{multirow}
\usepackage{tcolorbox}
\usepackage{hyperref}
\usepackage{url}
\usepackage[inline]{enumitem}
\usepackage{pifont}
\usepackage{tikz}
\usetikzlibrary{positioning,shapes.geometric}
\usepackage{pgfplots}
\pgfplotsset{compat=1.18}
\usepackage{wrapfig2}
\usepackage{needspace}
\usepackage[section]{placeins}
\usepackage{float}
\usepackage{algorithm}
\usepackage{algpseudocode}

\newtcolorbox{paperremark}{
  colback=black!2,
  colframe=black!70,
  boxrule=0.8pt,
  arc=2pt,
  left=6pt,
  right=6pt,
  top=5pt,
  bottom=5pt,
  before skip=6pt,
  after skip=6pt
}

\title{Learning Structured Sparsity for Large\\
Models: Bridging Model Compressibility\\
and Efficient Inference}

\author{
\textbf{Tianxiang Xia}\textsuperscript{1},
\textbf{Kai Lion}\textsuperscript{1},
\textbf{Niao He}\textsuperscript{1}, \\
\textbf{Michael Muehlebach}\textsuperscript{2,3},
\textbf{Melanie Zeilinger}\textsuperscript{1},
\textbf{Hao Ma}\textsuperscript{4}\ificlrfinal\thanks{Correspondence to: Hao Ma (\texttt{haoma@hku.hk}).}\else\ifpreprint\thanks{Correspondence to: Hao Ma.}\fi\fi
\ificlrfinal
\\[0.4em]
\textsuperscript{1}ETH Zurich, Zurich, Switzerland\\
\textsuperscript{2}Max Planck Institute for Intelligent Systems, T\"ubingen, Germany\\
\textsuperscript{3}Vienna University of Economics and Business, Vienna, Austria\\
\textsuperscript{4}The University of Hong Kong
\else
  \ifpreprint
\\{\normalfont\footnotesize\def\arraystretch{0.92}\begin{tabular}[t]{@{}l@{}}
\textsuperscript{1}ETH Z\"urich, Z\"urich \quad \textsuperscript{2}Max Planck Institute for Intelligent Systems, T\"ubingen\\
\textsuperscript{3}Vienna University of Economics and Business, Vienna \quad \textsuperscript{4}The University of Hong Kong\\
\texttt{\{xiatia, mzeilinger\}@ethz.ch, \{kai.lion, niao.he\}@inf.ethz.ch,}\\
\texttt{michael.muehlebach@wu.ac.at, haoma@hku.hk}
\end{tabular}}
  \fi
\fi
}

\newif\ifpreprint
\preprinttrue 
\newif\ifshowresources
\ificlrfinal\showresourcestrue\fi
\ifpreprint\showresourcestrue\fi
\ificlrfinal\else
  \ifpreprint
    \makeatletter
    \def\@maketitle{\vbox{\vskip -26pt\hsize\textwidth
      {\LARGE\sc \@title\par}
      \lhead{Preprint}
      \def\And{\end{tabular}\hfil\linebreak[0]\hfil
        \begin{tabular}[t]{l}\bf\rule{\z@}{24pt}\ignorespaces}%
      \def\AND{\end{tabular}\hfil\linebreak[4]\hfil
        \begin{tabular}[t]{l}\bf\rule{\z@}{24pt}\ignorespaces}%
      \begin{tabular}[t]{@{}l}\rule{\z@}{11pt}\ignorespaces\@author\end{tabular}%
      \par}}
    \makeatother
    \renewcommand{\iclrruler}[1]{}
  \fi
\fi
\begin{document}

\maketitle

\begin{abstract}
Scaling laws continue to reward increases in model scale, while the resulting computational and memory demands drive greater reliance on cloud infrastructure for inference. Yet unreliable network connectivity, communication delays, and privacy requirements can make such reliance impractical, motivating local execution on edge devices. Local deployment requires models to fit within limited memory and execute efficiently with the available computational resources. Recent work has shown that learning sparse-plus-low-rank representations enables elastic compression of a single checkpoint to multiple memory budgets. However, fewer parameters do not necessarily translate into faster inference. Indexing overhead and irregular memory access in unstructured sparse components can offset computational savings. To address this critical deployment barrier, we introduce S-SALAAD, a unified framework that jointly learns the model weights and their low-rank and structured sparse components during training. The framework supports a family of hardware-compatible structured sparsity patterns. It couples this structure learning with budget-aware compression and dedicated CPU and GPU kernels that exploit the learned structure for acceleration, enabling efficient inference from a single checkpoint across memory budgets. Experiments on Llama2 and Qwen3 models, covering both pretraining and distillation, show that S-SALAAD advances the empirical Pareto frontier in perplexity, memory, and inference latency across multiple hardware platforms relative to the evaluated baselines. These results show that structured learning can translate elastic compressibility into reduced memory requirements and lower inference latency, enabling practical deployment of large models on edge devices.
\end{abstract}


\section{Introduction}
\label{sec:intro}

Scaling laws~\citep{kaplan2020scaling} motivate larger models for improved quality, while their memory and computational demands increase reliance on cloud infrastructure with high-end accelerators.
However, network connectivity can be unstable in underwater and rescue operations~\citep{stojanovic2009underwater,deng2013task,gunn2015dynamic}.
Safety-critical applications such as autonomous driving require on-device decisions to avoid cloud communication delays~\citep{liu2019edge,kuutti2021survey,ma2025constraint,bodmer2026grounding}, while privacy requirements can prevent sensitive data from being uploaded~\citep{gdpr2016,shi2016edge}.
These constraints require large-model inference to move from the cloud to resource-constrained edge devices.
Fitting large models within limited memory and achieving low inference latency with the available computational resources are the two critical challenges of this transition.
The latency requirement is particularly important in feedback control, where inference latency adds to loop delay and excessive delays can degrade closed-loop performance or destabilize the system, compromising safety~\citep{shin1995computing,findeisen2004computational,ma2026stochastic,ma2023reinforcement,nan2026efficient}.

To address memory constraints, existing compression methods use quantization~\citep{dettmers2022llmint8}, distillation~\citep{gu2024minillm}, pruning~\citep{sun2024wanda,frantar2023sparsegpt,ashkboos2024slicegpt}, and low-rank factorization~\citep{hsu2022weightedlowrank,wang2025svdllm}.
Varying memory budgets across edge devices call for elastic compression of a single trained checkpoint to multiple model sizes.
SALAAD~\citep{ma2026salaad} enables such elastic compression by learning sparse-and-low-rank (SLR) representations for each linear projection\footnote{Here, \emph{linear projection} refers to a learned linear map in the attention (query/key/value/output) or multi-layer perceptron (MLP; up/gate/down) modules. Embeddings and the language model head are excluded unless noted.} during pretraining.
\textit{However, fewer parameters do not necessarily translate into lower inference latency.}
Unstructured SLR saves static memory but incurs higher inference latency (see Appendix~\ref{app:compression_latency_gap}).
We identify the overhead from irregular memory access in unstructured sparse operations as a critical barrier to deploying large models on edge devices.
Hardware-friendly patterns, including block sparsity~\citep{okanovic2025blast}, $N\!:\!M$ sparsity~\citep{zhao2024vnm}, and row-/column-wise sparsity~\citep{wen2016structured}, are therefore required to translate these memory savings into efficient edge inference.
A straightforward extension of SALAAD is to impose these patterns on its learned sparse components after pretraining.
However, our experiments show that this post-hoc conversion can sharply degrade model performance (see Appendix~\ref{app:salaad_as_blocks}).
We therefore induce \emph{structured sparsity} during \emph{pretraining}, allowing the SLR components to adapt jointly to the structural constraints.

To this end, we introduce \textbf{S-SALAAD} (\underline{S}tructured-\underline{S}parse-\underline{A}nd-\underline{L}ow-Rank \underline{A}daptation via \underline{AD}MM),
a plug-and-play framework that learns \textit{structured} SLR representations during pretraining for elastic compression and efficient edge inference.
Based on the alternating direction method of multipliers~(ADMM), it induces block, $N\!:\!M$, and row-/column-wise sparsity without changing the model architecture or optimizer update rules.
Efficient low-rank updates reduce pretraining overhead, while budget-aware refinement enables zero-shot compression of a single checkpoint to multiple parameter budgets without calibration data or further training.
Among these patterns, we identify block sparsity as a practical choice for elastic deployment given current technical capabilities.
With dedicated CPU and GPU kernels, the resulting models advance the perplexity--memory--latency Pareto frontier relative to the evaluated baselines.
Our contributions are summarized as follows:
\begin{itemize}[label=\ding{118}, leftmargin=1em, labelsep=0.5em, itemsep=0pt, parsep=0pt]
  \item \textbf{A plug-and-play framework for structured-sparsity learning.}
  S-SALAAD jointly learns model weights and their low-rank and structured sparse components under the task loss, adapting the representation to structural constraints during pretraining.
  S-SALAAD induces block, \(N\!:\!M\), and row-/column-wise sparsity without changing the model architecture, and it is compatible with existing optimizers.
  \item \textbf{A practical end-to-end implementation.}
  To support the training of larger models and their edge deployment, we address bottlenecks in the training, compression, and inference stages.
  Specifically, we use \begin{enumerate*}[label=\roman*), itemjoin={{; }}, itemjoin*={{; and }}]
    \item a warm-started iterative singular value decomposition~(I-SVD) to reduce the computational overhead of low-rank updates during pretraining
    \item energy-guided alternating refinement~(EGAR) to enable zero-shot compression without calibration data or further training, allocating the parameter budget across projections without a preset sparse--low-rank split and reducing reconstruction error within the allocated capacities
    \item dedicated CPU and GPU kernels to exploit the induced structured sparsity and reduce inference latency
  \end{enumerate*}.
  \item \textbf{Empirical Pareto frontier across training settings.}
  S-SALAAD advances the empirical perplexity--memory--latency Pareto frontier across hardware platforms relative to the evaluated baselines, with each checkpoint supporting multiple compression budgets.
  These gains hold for Llama2-1B pretraining and cross-architecture distillation from Nemotron-3-Super-120B to Qwen3-1.7B.
  \item
  \textbf{Analysis of structured representations.}
  S-SALAAD provides an explicit structured representation of Transformer weights, enabling us to identify distinct sparsity characteristics across linear projection types.
  Our analysis identifies block sparsity as a practical choice for advancing the perplexity-memory-latency Pareto frontier, combining flexible compression with efficient execution on CPUs and GPUs.
  Furthermore, we also reveal the relationship between tolerable block granularity and model scale, providing empirical guidance for selecting block sizes when extending S-SALAAD to larger models.
\end{itemize}

\section{Related Work}
\label{sec:related}

Compression methods for large language models broadly divide into pruning, low-rank factorization, and quantization.
Unstructured pruning removes individual weights by magnitude \citep{han2015learning} or calibration-based criteria \citep{sun2024wanda,frantar2023sparsegpt}, while low-rank methods such as SVD-LLM \citep{wang2025svdllm} factorize weight matrices and SliceGPT \citep{ashkboos2024slicegpt} shrinks them by removing channels.
A second post-hoc family instead runs iterative ADMM on the pretrained weights: ELSA \citep{lee2026elsa} pushes unstructured sparsity to extreme ratios, and 3BASiL \citep{makni20253basil} performs layer-wise SLR reconstruction via a 3-block ADMM--both are data-hungry and take hours of GPU-time to compress a single model.
Compression-aware pretraining instead induces compressibility during training: LORO \citep{mo2025loro} pretrains in a preset low-rank subspace, SLTrain \citep{han2024sltrain} augments low-rank pretraining with a fixed-support sparse component, and SALAAD \citep{ma2026salaad} jointly learns SLR components via ADMM.
The sparse components in SLTrain and SALAAD are unstructured and do not map efficiently onto accelerator primitives.
LOST~\citep{li2025lost} combines low-rank factors with channel-wise sparsity, but its preset rank and sparse capacity fix the model size during training, and its formulation does not support elastic deployment across budgets from a single checkpoint.
It also inserts an additional nonlinearity between the low-rank factors, changing the underlying model architecture.
S-SALAAD preserves the original architecture and supports multiple structured sparsity patterns, learning a representation that can be compressed from a single checkpoint to different parameter budgets.


\section{Problem Formulation}
\label{sec:problem}

Let $\{X_i\}_{i=1}^{E}$ denote the weight matrices of a neural network's $E$ target linear projections, where $X_i\!\in\!\R^{n_i \times m_i}$.
For notational simplicity, we omit the projection index $i$ below.
SALAAD~\citep{ma2026salaad} decomposes each projection matrix $X$ into a low-rank component $L\!\in\!\R^{n \times m}$ and an unstructured element-wise sparse component $S\!\in\!\R^{n \times m}$ by solving the following optimization problem:
\begin{equation}
  \min_{X,L,S} \; \ell(X) + \alpha \left|L\right|_* + \beta \left|S\right|_1 \quad \text{s.t.} \; X = L + S,
  \label{eq:salaad}
\end{equation}
where $\ell: \mathbb{R}^{n \times m}\!\rightarrow\!\mathbb{R}$ denotes the task-specific loss.
The nuclear norm $\left|\cdot\right|_{*}$ and $\ell_1$-norm $\left|\cdot\right|_{1}$ serve as convex surrogates for matrix rank and the number of nonzero entries, respectively, while $\alpha\!>\!0$ and $\beta\!>\!0$ control their regularization strengths.

To induce hardware-friendly structure in $S$, we reformulate~\eqref{eq:salaad} as follows:
\begin{equation}
  \min_{X,L,S} \; \ell(X) + \alpha \left|L\right|_* + \underbrace{\sum_{d=1}^D \beta_d \Pi(S^d)}_{\Omega\left(S\right)} \quad \text{s.t.} \; X = L + S,
  \label{eq:s-salaad}
\end{equation}
where $\Omega(S)$ denotes the structured sparsity regularizer.
More specifically, we partition $S$ into $D$ non-overlapping groups, with $S^d$ denoting the $d$-th group.
Let $\Pi(\cdot)$ denote the norm applied to each group, with $\beta_d\!>\!0$ controlling the regularization strength of group $d$.
The partition and the norm assigned to each group jointly determine the sparsity structure promoted by $\Omega(S)$ in~\eqref{eq:s-salaad}.
Here, $X$, $L$, and $S$ are jointly optimized under the task loss, allowing the model weights and both components to adapt to the prescribed sparsity structure during pretraining.
The element-wise sparsity induced in SALAAD is recovered by partitioning $S\!\in\!\R^{n\times m}$ into $D\!=\!nm$ singleton groups, assigning the absolute-value norm to every group, and setting $\beta_d\!=\!\beta$ for all $d\!\in\!\left\{1,\dots,D\right\}$.
This instantiation gives $\Omega(S)\!=\!\beta\sum_{i=1}^{n}\sum_{j=1}^{m}|S_{ij}|\!=\!\beta |S|_1$, making the unstructured sparsity a special case of \eqref{eq:s-salaad}.

We next present three instantiations that induce sparsity patterns widely regarded as hardware-friendly.
These choices are not exhaustive, and the framework can be instantiated to induce other structures as hardware capabilities and deployment requirements evolve.

\textbf{Block sparsity.}
Assuming that $p, q\!\in\!\mathbb{Z}_{+}$ divide $n$ and $m$ respectively, we partition $S$ into $D\!=\!\nicefrac{nm}{pq}$ non-overlapping blocks $\{S^d\}_{d=1}^{D}$ of size $p \times q$.
Setting $\Pi(S^d)\!=\!|S^d|_\mathrm{F}$ and $\beta_d\!=\!\beta$ for every block gives
\begin{equation}
  \Omega \left(S\right) \doteq \beta\sum_{d=1}^{D} \left|S^d\right|_{\mathrm{F}}.
  \label{eq:block_penalty}
\end{equation}
Each block $S^d$ is treated as a single group, whose magnitude is measured by the Frobenius norm $|\cdot|_{\mathrm{F}}$.
Penalizing the sum of these block magnitudes removes a block as a whole rather than pruning its entries independently, thereby inducing block sparsity.
This regularizer follows the group-LASSO construction of~\citet{yuan2006group}, with each block treated as a group.

\textbf{Row-/column-wise sparsity.}
Row- and column-wise sparsity are obtained from block sparsity by setting $(p,q)\!=\!(1,m)$ and $(p,q)\!=\!(n,1)$, respectively.
The resulting groups are the $D\!=\!n$ rows or $D\!=\!m$ columns of $S$, respectively, with $\Pi(S^d)\!=\!|S^d|_2$.
Setting $\beta_d\!=\!\beta$ for every group gives
\begin{equation}
    \Omega_{\mathrm{row}}(S) \doteq \beta\sum_{d=1}^{D}\left|S^d\right|_2 = \beta\sum_{d=1}^{n}\left|S_{d,:}\right|_2, \quad 
    \Omega_{\mathrm{col}}(S) \doteq \beta\sum_{d=1}^{D}\left|S^d\right|_2 = \beta\sum_{d=1}^{m}\left|S_{:,d}\right|_2.
  \label{eq:row_col_penalty}
\end{equation}
The resulting group norms induce complete row or column removal.

\textbf{$N\!\!:\!\!M$ sparsity.}
Assuming that $M\!\in\!\mathbb{Z}_{+}$ divides $m$, we partition every row of $S$ into consecutive groups of $M$ entries, yielding $D\!=\!\nicefrac{nm}{M}$ non-overlapping groups $\{S^d\}_{d=1}^{D}$.
Setting $\Pi(S^d)\!=\!|S^d|_1$ gives
\begin{equation}
  \Omega \left(S\right) \doteq \sum_{d=1}^{D}\beta_d \left|S^d\right|_1,
  \label{eq:nm_penalty}
\end{equation}
where $\beta_d$ specifies the regularization strength of the $d$-th group.
The target of retaining $N$ nonzero entries in every group of $M$ entries defines $N\!:\!M$ sparsity~\citep{zhao2024vnm}.

\textbf{$L$-only.}
Fixing $S\equiv0$ in~\eqref{eq:s-salaad} yields the $L$-only special case with $X\!=\!L$.
We use it as a reference for tolerable block granularity across model scales; further discussion appears in Appendix~\ref{sec:ablation}.

\section{Methodology}
\label{sec:method}

This section describes how S-SALAAD jointly learns the model weights and their low-rank and structured sparse components during pretraining and supports compression and efficient inference for edge deployment.

\subsection{Structured SLR Learning}
\label{sec:algorithm}

We use ADMM~\citep{hong2016convergence,candes2011robust,lin2010augmented,ouyang2013stochastic} to jointly optimize the model weights and their structured SLR components in~\eqref{eq:s-salaad}.
Introducing a dual variable $Y\!\in\!\R^{n \times m}$ and a penalty parameter $\rho\!>\!0$ gives the augmented Lagrangian
\begin{equation}
  \gG_\rho \left(X,L,S,Y\right) = \ell \left(X\right) + \alpha\left|L\right|_* + \Omega\left(S\right)
  + \frac{\rho}{2} \Big| X-L-S + \frac{Y}{\rho} \Big|_\mathrm{F}^2
  - \frac{1}{2\rho} \left|Y\right|_{\mathrm{F}}^2.
  \label{eq:auglag}
\end{equation}
We describe the selection of $\rho$ in Appendix~\ref{app:hparams}.
Within this alternating optimization, the choice of $\Omega(S)$ determines the structured-sparsity update.
At each iteration $k\!\geq\!1$, ADMM updates the primal variables $(X_k,L_k,S_k)$ and the dual variable $Y_k$ as follows.

\textbf{Model update.}
Because the model-update subproblem $\argmin_X \gG_\rho(X,L_{k-1},S_{k-1},Y_{k-1})$ does not admit a closed-form solution, we approximate its solution $X_k$ through gradient-based optimization.
Any existing optimizer can be used without modification, with the original training loss $\ell(X)$ replaced by the coupled loss
\begin{equation}
  \ell_c (X)
  \doteq  \ell(X)
  + \frac{\rho}{2} \Big| X-L_{k-1}-S_{k-1}+\frac{Y_{k-1}}{\rho} \Big|_\mathrm{F}^2.
  \label{eq:coupled_loss}
\end{equation}

\textbf{Low-rank proximal update.}
The low-rank update is given by
\begin{equation}
  L_k = \mathrm{prox}_{\frac{\alpha}{\rho}\left|\cdot\right|_{*}} (\underbrace{X_k - S_{k-1} + \nicefrac{Y_{k-1}}{\rho}}_{\overset{\mathrm{SVD}}{=} U_k\,\mathrm{diag}(\sigma_k)V_k^\top}) = U_k\,\mathrm{diag} \big(\{\sigma_{k,j}-\nicefrac{\alpha}{\rho}\}_+\big)V_k^\top,
  \label{eq:lstep}
\end{equation}
where $\sigma_k\!=\!\{\sigma_{k,j}\}_j$ collects the singular values at iteration $k$, and $\{\cdot\}_+$ denotes the element-wise positive part.

\textbf{Structured-sparsity proximal update.}
With $Z_k\!\doteq\!X_k-L_k+\nicefrac{Y_{k-1}}{\rho}$, the sparse update can be written as the following proximal problem:
\begin{equation}
  S_k
  =\argmin_S \frac{1}{\rho}\Omega(S)
  +\frac{1}{2}\left|S-Z_k\right|_\mathrm{F}^2
  =\operatorname{prox}_{\frac{1}{\rho}\Omega(S)}(Z_k).
  \label{eq:s_subproblem}
\end{equation}
The non-overlapping partition underlying $\Omega(S)\!=\!\sum_{d=1}^{D}\beta_d\Pi(S^d)$ decomposes this proximal problem into $D$ independent group-wise subproblems.
For each $d\!\in\!\{1,\ldots,D\}$, let $Z_k^{d}$ and $S_k^{d}$ denote the components of $Z_k$ and $S_k$ associated with the $d$-th group, respectively.
Under the block-sparsity regularizer and its row-/column-wise specializations, the uniform choice $\beta_d\!=\!\beta$ yields the group soft-thresholding update
\begin{equation}
  S_k^{d}
  =\frac{1}{\rho} \Big\{ \rho - \frac{\beta}{|Z_k^{d}|_\mathrm{F}}\Big\}_+ Z_k^{d}, \quad S_k^{d} \doteq 0 \text{ if } Z_k^{d}=0.
  \label{eq:block_prox}
\end{equation}
Under the $N\!:\!M$ regularizer, the group-specific coefficients $\{\beta_d\}_{d=1}^{D}$ yield the element-wise soft-thresholding update within each group
\begin{equation}
  S_k^{d}
  =\sign\!\big(Z_k^{d}\big) \odot
  \Big\{\big|Z_k^{d}\big|-\frac{\beta_d}{\rho}\Big\}_{+},
  \label{eq:nm_prox}
\end{equation}
where $\odot$ denotes element-wise multiplication.

\textbf{Dual ascent.} The dual variable is updated by gradient ascent on the Lagrangian $Y_k\!=\!Y_{k-1}\!+\!\rho\left(X_k\!-\!L_k\!-\!S_k\right)$.

\subsection{End-to-End Implementation}
\label{sec:implementation}

\begin{wrapfloat}{algorithm}[-3]{R}{0.5\linewidth}*
  \vspace*{-\intextsep}
  \begin{minipage}[t]{\linewidth}
  \vspace{0pt}
  \captionsetup{labelfont=bf,labelsep=space,justification=raggedright,singlelinecheck=false,skip=0pt}
  \hrule height 0.8pt depth 0pt
  \kern2pt
  \caption{S-SALAAD Training}
  \label{alg:ssalaad_pretrain}
  \kern2pt
  \hrule
  \kern2pt
  \scriptsize
  \begin{algorithmic}
  \State \textbf{Initialization:} $X, L, S, Y, V$, and structured sparsity $\Omega(S)$
  \State \textbf{Input:} $\alpha, \beta$, and $\rho$
  \While{True}
    \State $X \gets \textsc{Backpropagation on } \ell_c (X)$ over minibatches \hfill \eqref{eq:coupled_loss}
    \State $L_0 \gets L$, $S_0 \gets S$, $Y_0 \gets Y$, $V_0 \gets V$
    \State \textcolor{red}{\textit{/* \begin{sc}Structured SLR Learning\end{sc} */}}
    \For{$k = 1, \dots, K$}
    \State $W_k \doteq X - S_{k-1} + \nicefrac{Y_{k-1}}{\rho}$
    \State $U_k, \sigma_k, V_k\gets \textsc{I-SVD-Step}(W_k, V_{k-1})$ \hfill Alg.~\ref{alg:iterative_svd}
    \State $L_k \gets \textsc{L-Step}(U_k, \sigma_k, V_k;\, \alpha, \rho)$ \hfill \eqref{eq:lstep}
    \State $Z_k \doteq X - L_k + \nicefrac{Y_{k-1}}{\rho}$
    \State $S_k \gets \textsc{S-step}(Z_k;\, \beta, \rho, \Omega)$ \hfill \eqref{eq:block_prox} or\,\eqref{eq:nm_prox}
    \State $Y_k = Y_{k-1} + \rho (X - L_k - S_k)$
    \EndFor
    \State $L \gets L_K$, $S \gets S_K$, $Y \gets Y_K$, $V \gets V_K$
    \State $\alpha, \beta \gets \textsc{I-Controller}(\alpha, \beta, L, S)$
  \EndWhile
  \end{algorithmic}
  \kern2pt
  \hrule
\end{minipage}
  \par\vspace{0.5\baselineskip}
  \begin{minipage}[t]{\linewidth}
  \vspace{0pt}
  \captionsetup{type=algorithm,labelfont=bf,labelsep=space,justification=raggedright,singlelinecheck=false,skip=0pt}
  \hrule height 0.8pt depth 0pt
  \kern2pt
  \caption{Iterative SVD Step}
  \label{alg:iterative_svd}
  \kern2pt
  \hrule
  \kern2pt
  \scriptsize
  \begin{algorithmic}
  \Statex \textbf{Input}\quad $W$, cached basis $V$ with $m$ rows
  \State $(V,R) \gets \textsc{QR}\left(W^\top ( W V )\right)$
  \State $\sigma \gets \sqrt{\mathbf{1}^\top ((W V) \odot (W V))}$ \Comment{$\mathbf{1}$: all-ones column vector}
  \State $U \gets (W V)\,\mathrm{diag}(\sigma)^{-1}$
  \State \textbf{Return}\quad $(U,\sigma,V)$
  \end{algorithmic}
  \kern2pt
  \hrule
  \end{minipage}
  \par\vspace{0.5\baselineskip}
  \begin{minipage}[t]{\linewidth}
  \centering
  \captionsetup{type=figure}
  \includegraphics[width=\linewidth]{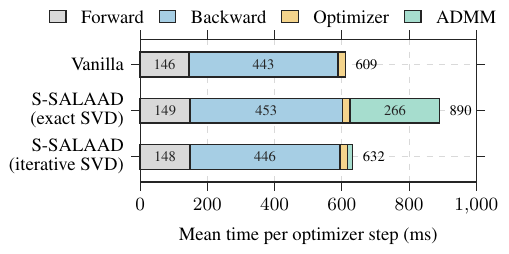}
  \caption{Per-step training-time breakdown for Llama-1B on $8$ H100 GPUs.}
  \label{fig:svdt_training_breakdown}
  \end{minipage}
\end{wrapfloat}

To support large-scale training and elastic edge deployment, we address the computational cost of low-rank updates during pretraining, the zero-shot budget-aware compression of the learned components, and efficient inference on CPUs and GPUs.

\textbf{Training with I-SVD.} Algorithm~\ref{alg:ssalaad_pretrain} summarizes the complete training procedure, which alternates model updates with structured SLR learning.
For training efficiency, we update $L$, $S$, and $Y$ once per structured SLR learning phase.
Details of the I-Controller are provided in Appendix~\ref{app:i_controller}.

In Algorithm~\ref{alg:ssalaad_pretrain}, the low-rank update in~\eqref{eq:lstep} applies singular-value soft thresholding to $X_k\!-\!S_{k-1}\!+\!\nicefrac{Y_{k-1}}{\rho}$ at iteration $k\!\geq\!1$.
Computing an exact SVD at every iteration introduces substantial training overhead as shown in \figref{fig:svdt_training_breakdown}.

We therefore use I-SVD (see Algorithm~\ref{alg:iterative_svd}), which performs a single step of orthogonal iteration~\citep{golub2013matrix} warm-started from the cached right singular basis $V_{k-1}$.
It returns approximate triplets $(U_k,\sigma_k,V_k)$ and caches $V_k$ for the next iteration.

The step applies $W_k^\top W_k$ to the cached basis $V_{k-1}$ and orthogonalizes the product by QR, giving the new basis $V_k$, and the triangular factor is discarded.
From the single product $W_kV_k\!=\!U_k\,\mathrm{diag}(\sigma_k)$, we obtain $\sigma_k$ as the column norms and $U_k$ as the normalized columns.
Warm-starting from the previous basis enables efficient tracking across successive updates.
Empirically, a single step of orthogonal iteration preserves compression quality and closely tracks the effective-rank trajectories obtained with exact SVD, as shown in Appendix~\ref{app:iterative_svd_results}.

\figref{fig:svdt_training_breakdown} quantifies the resulting efficiency gain.
Compared with exact SVD, iterative SVD reduces the mean time per optimizer step from \qty{890}{\milli\second} to \qty{632}{\milli\second}, a \qty{29}{\percent} reduction.
Relative to the \qty{609}{\milli\second} vanilla baseline, this leaves only \qty{4}{\percent} overhead.
Thus, I-SVD nearly removes the ADMM overhead in S-SALAAD, making end-to-end training practical at scale.

\textbf{Energy-guided alternating refinement for budget-aware compression.}
To enable elastic deployment, we need to compress a trained checkpoint to different parameter budgets without calibration data or further training.
Given a target budget $P_{\mathrm{tgt}}$, \emph{energy-guided alternating refinement} (EGAR) first allocates low-rank and sparse capacities across projections, then refines both components within these capacities to improve reconstruction of the dense weights $\{X_i\}_{i=1}^E$.
Both stages preserve the sparsity structure induced during pretraining.
We describe block sparsity below; row-/column-wise sparsity follows by changing the partition.\footnote{For $N\!:\!M$ sparsity, Stage~I keeps the fixed-capacity $S$ and allocates only $N_L$; Stage~II instead uses $\mathcal{R}_S \doteq \{A \mid |A^d|_0\!\leq\!N,\ d=1,\ldots,D\}$, where $A^d$ is the $d$-th group of $M$ entries.}

\begin{wrapfigure}[15]{r}{0.34\linewidth}
  \vspace{-\baselineskip}
  \centering
  \includegraphics[width=\linewidth]{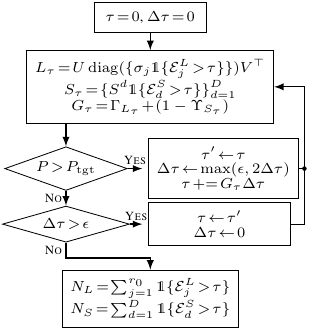}
  \caption{Pipeline of Stage I.}
  \label{fig:egar_budget_allocation}
  \vspace{-\baselineskip}
\end{wrapfigure}
\textit{Stage I: Energy-guided budget allocation.}
To account for heterogeneity across linear projections, Stage~I adaptively allocates the global parameter budget $P_{\mathrm{tgt}}$ across projections and between their low-rank and sparse components.
The resulting retained ranks and sparse-block counts define the capacity constraints for Stage~II.

Allocation is guided by the energy per stored parameter.
Omitting the projection index, we score singular directions and sparse blocks as
\begin{equation*}
  \label{eq:scores}
    \mathcal{E}^L_j\!\doteq\!\nicefrac{\sigma_j^2}{n+m},~j\!=\!1,\ldots,r_0,~
    \mathcal{E}^S_d\!\doteq\!\nicefrac{|S^d|_{\mathrm{F}}^2}{pq},~d\!=\!1,\ldots,D,
\end{equation*}
where $\{\sigma_j\}$ are the singular values of $L$, $r_0\!=\!\min(n,m)$, and $\{S^d\}_{d=1}^{D}$ partitions $S$ into $p\times q$ blocks.
Each projection has its own threshold $\tau$, shared by its low-rank and sparse components; candidates with scores greater than $\tau$ are retained.

For a feasible budget $P_{\mathrm{tgt}}$, the coarse-to-fine search in~\figref{fig:egar_budget_allocation} adjusts the threshold $\tau$ using a minimum base step $\epsilon\!>\!0$.
Threshold increments are scaled by $G_\tau\!=\!\Gamma_{L_\tau}\!+\!(1\!-\!\Upsilon_{S_\tau})$, using the remaining-capacity statistics defined in Appendix~\ref{app:allocation_statistics}.
The search returns the retained direction and block counts $(N_L,N_S)$ for each projection, with total model parameter count $P\!\leq\!P_{\mathrm{tgt}}$.

\textit{Stage II: Alternating refinement.}
Thresholding in Stage~I can increase reconstruction error.
Using the dense weight $X$ as the reconstruction anchor, Stage~II alternately refines the low-rank and sparse components $L$ and $S$ within the allocated capacities to reduce the reconstruction error while preserving the parameter budget.
We initialize $(L,S)\!=\!(L_\tau,S_\tau)$ and define the feasible sets
\begin{equation*}
  \mathcal{R}_L\!\doteq\!
  \bigl\{A \in\mathbb{R}^{n\times m} \mid \operatorname{rank}(A)\leq N_L\bigr\},~
  \mathcal{R}_S\!\doteq\!
  \bigl\{A \in\mathbb{R}^{n\times m}\mid A\text{ has at most }N_S\text{ nonzero blocks}\bigr\}.
\end{equation*}
We then alternate the residual projections
\begin{equation}
  L \leftarrow \operatorname{Proj}_{\mathcal{R}_L}(X-S),
  \quad
  S \leftarrow \operatorname{Proj}_{\mathcal{R}_S}(X-L).
  \label{eq:refinement}
\end{equation}
The projections retain, respectively, the leading $N_L$ singular components and the $N_S$ blocks with the largest Frobenius norms of the corresponding residuals.
Each update minimizes $|X-L-S|_{\mathrm{F}}^2$ over its feasible set with the other component fixed, so reconstruction error is non-increasing (see Appendix~\ref{app:energy_guided_alternating_refinement}).
The low-rank subspace and block support remain free to change within the allocated capacities.
Further details and empirical results are provided in Appendix~\ref{app:energy_guided_alternating_refinement}.

\textbf{Dedicated CPU and GPU kernels.}
Among the three supported sparsity patterns, we identify block sparsity as a practical choice for elastic edge deployment, as discussed in Section~\ref{sec:ablation_studies}.
We therefore develop dedicated CPU and GPU kernels only for block-sparse inference.
At inference, the compressed linear map $L\!+\!S$, with $L\!=\!AB^\top$, is applied to an input activation $h\!\in\!\mathbb{R}^{m}$ as
\begin{equation}
  y=(L+S)h=(AB^\top+S)h=A(B^\top h)+Sh.
  \label{eq:inference}
\end{equation}
We first compute $y\!=\!AB^\top h$ using dense kernels, and then accumulate the sparse contribution $Sh$ directly into $y$, avoiding a separate output buffer for the sparse path.
For the sparse path, we store only the retained blocks in a contiguous layout with block-level row pointers and column indices, amortizing indexing overhead across the weights within each block.
The kernels skip zero blocks and exploit contiguous access and regular computation within each retained block to reduce inference latency.
We implement the GPU kernel in TileLang~\citep{wang2025tilelang} to perform dense tile computations on Tensor Cores, and vectorize the CPU kernel with AVX2.
Kernel comparisons are reported in Appendix~\ref{app:kernel_speedup}.

\section{Experiments}
\label{sec:experiments}

We pretrain Llama2 models (1B and 350M) on C4~\citep{raffel2020c4} and distill Nemotron-3-Super-120B into Qwen3-1.7B on FineWeb-Edu~\citep{penedo2024fineweb}.
Dataset and model details are provided in Appendix~\ref{app:datasets}, and training settings in Appendix~\ref{app:hparams}. We compare against Wanda~\citep{sun2024wanda}, SparseGPT~\citep{frantar2023sparsegpt}, SliceGPT~\citep{ashkboos2024slicegpt}, SVD-LLM~\citep{wang2025svdllm}, LORO~\citep{mo2025loro}, vanilla training, and SALAAD~\citep{ma2026salaad}.
LOST~\citep{li2025lost} is excluded because it requires changes to the underlying model architecture.
We evaluate perplexity, zero-shot downstream accuracy, peak memory, and prefill inference latency.
To assess deployment under different memory and compute constraints, we profile inference on Quadro~RTX~6000 and RTX~2080~Ti GPUs and 4$\times$ AMD~EPYC~7662 CPU hosts.
Baseline implementations and evaluation protocols are detailed in Appendices~\ref{app:evaluation} and~\ref{app:compute}.

\subsection{Main Results}
\label{sec:results}

\begin{wrapfigure}{R}{0.67\textwidth}
  \vspace{-0.5em}
  \centering
  \setlength{\parskip}{0pt}
  \captionsetup{type=figure,skip=4pt}
  \captionsetup[subfigure]{font=small,skip=1pt,justification=centering}
  \newlength{\figthreeleftwidth}
  \newlength{\figthreerightwidth}
  \setlength{\figthreeleftwidth}{0.32115\linewidth}
  \setlength{\figthreerightwidth}{\dimexpr\linewidth-\figthreeleftwidth\relax}
  \begin{subfigure}[t]{\figthreeleftwidth}
    \centering
    \includegraphics[width=\linewidth]{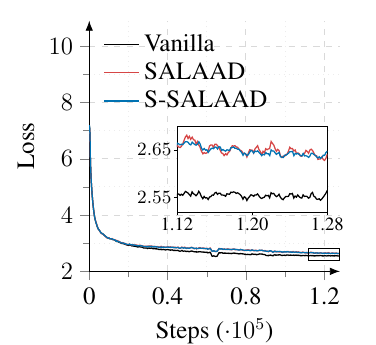}
    \caption{Training loss.}
    \label{fig:structure_training_loss}
  \end{subfigure}%
  \begin{subfigure}[t]{\figthreerightwidth}
    \centering
    \includegraphics[width=\linewidth]{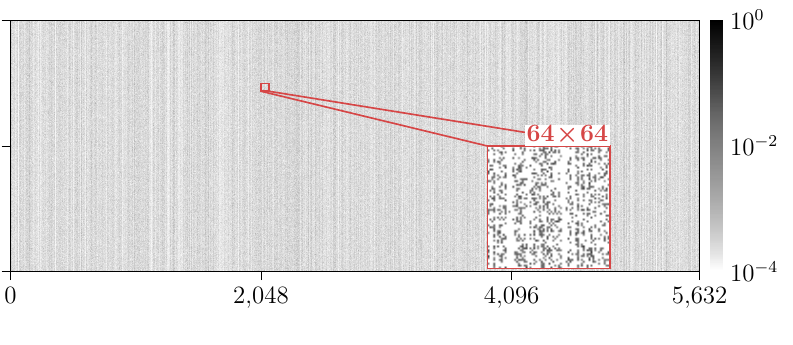}
    \caption{Down projection.}
    \label{fig:structure_unstructured_heatmap}
  \end{subfigure}

  \vspace{3pt}
  \begin{subfigure}[t]{\figthreeleftwidth}
    \centering
    \includegraphics[width=\linewidth]{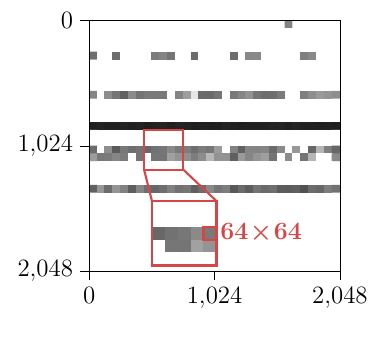}
    \caption{Q projection.}
    \label{fig:structure_block_heatmap_qproj}
  \end{subfigure}%
  \begin{subfigure}[t]{\figthreerightwidth}
    \centering
    \includegraphics[width=\linewidth]{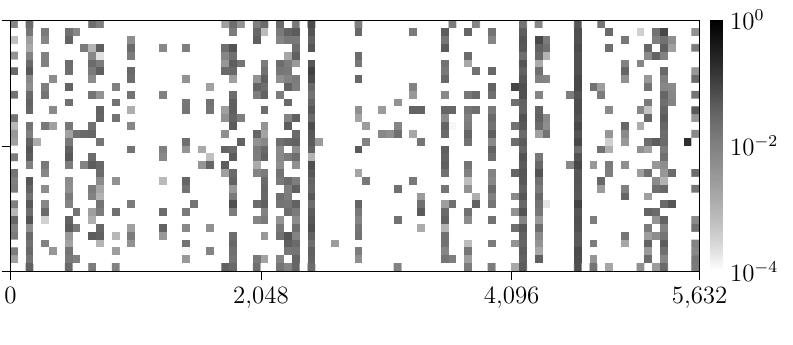}
    \caption{Down projection.}
    \label{fig:structure_block_heatmap}
  \end{subfigure}
  \caption{Llama-1B training and learned $S$ in a randomly selected layer. Log shading shows entry magnitudes for SALAAD (\subref{fig:structure_unstructured_heatmap}) and $64\!\times\!64$ block Frobenius norms for S-SALAAD (\subref{fig:structure_block_heatmap_qproj},\subref{fig:structure_block_heatmap}), clipped to the displayed range.}
  \label{fig:dynamics_and_heatmaps}
\end{wrapfigure}

\begin{figure}[t]
  \centering
  \begin{subfigure}[b]{0.49\linewidth}
    \centering
    \includegraphics[width=\linewidth]{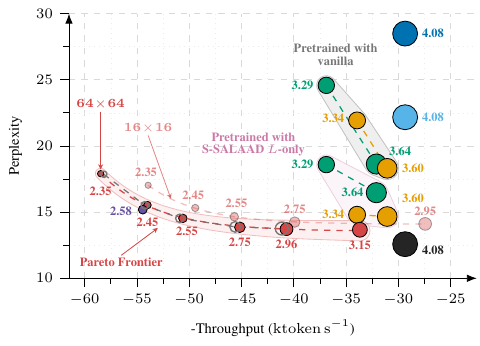}
    \caption{Llama-1B.}
    \label{fig:e2e_gpu}
  \end{subfigure}\hfill
  \begin{subfigure}[b]{0.49\linewidth}
    \centering
    \includegraphics[width=\linewidth]{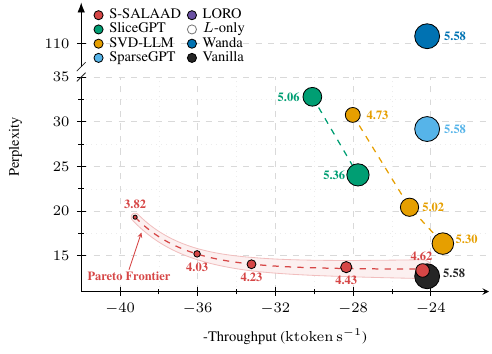}
    \caption{Qwen3-1.7B.}
    \label{fig:baselines_qwen}
  \end{subfigure}
  \caption{Perplexity against throughput on a Quadro~RTX~6000; marker radius linearly encodes peak GPU memory usage, with values annotated in GB. Lighter red denotes the $16\!\times\!16$ variant.}
  \label{fig:e2e}
\end{figure}

\textbf{Training stability and structure induction.}
Figure~\ref{fig:structure_training_loss} shows that S-SALAAD ($64\!\times\!64$) on Llama-1B closely follows the loss trajectory of SALAAD, with slightly higher loss than vanilla.
This modest loss penalty is accompanied by substantial memory savings and lower inference latency after compression, yielding the favorable deployment trade-offs reported in Figure~\ref{fig:e2e}.
Stable training is also observed across sparsity patterns on Llama-350M (Figure~\ref{fig:training_350m_loss}), block granularities on Llama-1B (Figure~\ref{fig:training_1b_loss}), and Qwen3-1.7B distillation (Figure~\ref{fig:training_qwen_loss}).
Figures~\ref{fig:structure_unstructured_heatmap} and~\ref{fig:structure_block_heatmap} contrast element-wise and block sparsity in the down projection of the same randomly selected layer; Figure~\ref{fig:structure_block_heatmap_qproj} shows the structured query projection.
Figures~\ref{fig:training_350m_heatmaps}--\ref{fig:training_qwen_heatmaps} provide additional heatmaps across these settings; Figures~\ref{fig:training_350m_rank_sparsity}--\ref{fig:training_qwen_rank_sparsity} show effective-rank and sparsity trajectories illustrating progressive structure induction under the prescribed targets.

\textbf{Pareto dominance over post-training baselines.}
\figref{fig:e2e} compares perplexity, throughput, and peak GPU memory jointly on a Quadro~RTX~6000, taking the uncompressed model obtained through standard training as the vanilla-pretrained reference.
Under conventional Llama pretraining, S-SALAAD establishes a superior Pareto frontier over Wanda, SparseGPT, SliceGPT, and SVD-LLM (see \figref{fig:e2e_gpu}).

Indeed LORO also lies on this Pareto frontier, although each of its checkpoints is tied to a fixed model size, whereas a single S-SALAAD checkpoint supports elastic deployment across multiple parameter budgets.
At the 500M operating point, block-$64\!\times\!64$ S-SALAAD reduces peak GPU memory from \qty{4.08}{\giga\byte} to \qty{2.35}{\giga\byte} (\qty{42.4}{\percent}) and increases throughput from \qty{29.35}{\kilo\token\per\second} to \qty{58.47}{\kilo\token\per\second} ($1.99\times$) relative to dense model inference.
Plus, in compression, Wanda, SparseGPT, SliceGPT, and SVD-LLM require calibration samples to estimate statistics, whereas EGAR determines the compressed $L\!+\!S$ representation solely from the learned $L$, $S$, and dense weight $X$.
The same advantage persists under cross-architecture distillation (\figref{fig:baselines_qwen}): at 800M, S-SALAAD reduces peak GPU memory from \qty{5.58}{\giga\byte} to \qty{3.82}{\giga\byte} (\qty{31.5}{\percent}) and increases throughput from \qty{24.19}{\kilo\token\per\second} to \qty{39.22}{\kilo\token\per\second} ($1.62\times$).
Together, these results show that the Pareto-frontier gains extend across model families and training paradigms.

Applying SliceGPT and SVD-LLM to an S-SALAAD-pretrained checkpoint ($L$-only) further improves their trade-offs relative to their vanilla-pretrained counterparts, highlighting the \textit{synergies} between S-SALAAD's learned compressibility and post-training compression methods.
Additional evaluations on an RTX~2080~Ti and 4$\times$ AMD~EPYC~7662 CPUs are reported in Appendix~\ref{app:additional_hardware}.

\textbf{Downstream evaluation.}
Table~\ref{tab:downstream} compares zero-shot downstream accuracy between vanilla models and compressed variants across different parameter budgets, using the same Llama and Qwen checkpoints trained with S-SALAAD as in the Pareto-frontier evaluation (see Figure~\ref{fig:e2e}). Across both settings, accuracy degrades gracefully rather than collapsing as the parameter budget tightens. This trend demonstrates both that a single S-SALAAD checkpoint supports elastic deployment across capacities and that training-time induction of block-sparse SLR structure preserves the model's downstream capability.
\begin{table}[!htbp]
  \centering
  \caption{Zero-shot accuracy of Vanilla and S-SALAAD (in blue) across budgets.}
  \label{tab:downstream}
  \normalsize
  \setlength{\tabcolsep}{2.5pt}
  \renewcommand{\arraystretch}{1}
\definecolor{paretoRowBlue}{HTML}{EAF3FF}
\begin{tabularx}{\linewidth}{@{}l>{\raggedleft\arraybackslash}X*{4}{>{\columncolor{paretoRowBlue}\raggedleft\arraybackslash}X}@{\hspace{8pt}}>{\raggedleft\arraybackslash}X*{4}{>{\columncolor{paretoRowBlue}\raggedleft\arraybackslash}X}}
\toprule
\shortstack[l]{} & \multicolumn{5}{c}{Llama-1B} & \multicolumn{5}{c}{Qwen3-1.7B} \\
\cmidrule(lr){2-6}\cmidrule(lr){7-11}
Task & Vanilla & 800M & 700M & 600M & 500M & Vanilla & 1273M & 1200M & 1000M & 800M \\
\midrule
PIQA & 0.72 & 0.71 & 0.70 & 0.69 & 0.68 & 0.73 & 0.72 & 0.72 & 0.72 & 0.70 \\
SciQ & 0.79 & 0.76 & 0.76 & 0.76 & 0.77 & 0.92 & 0.92 & 0.91 & 0.90 & 0.89 \\
WinoG. & 0.54 & 0.54 & 0.54 & 0.54 & 0.54 & 0.57 & 0.58 & 0.59 & 0.58 & 0.56 \\
HellaSwag & 0.39 & 0.37 & 0.37 & 0.37 & 0.35 & 0.46 & 0.45 & 0.44 & 0.43 & 0.40 \\
ARC-Easy & 0.54 & 0.51 & 0.51 & 0.51 & 0.50 & 0.76 & 0.74 & 0.73 & 0.72 & 0.68 \\
OBQA & 0.23 & 0.20 & 0.20 & 0.20 & 0.19 & 0.30 & 0.28 & 0.27 & 0.27 & 0.26 \\
\bottomrule
\end{tabularx}

\end{table}

\subsection{Ablation Studies}
\label{sec:ablation_studies}
\begin{wrapfigure}[12]{R}{0.34\textwidth}
  \vspace{-0.5em}
  \centering
  \includegraphics[width=\linewidth]{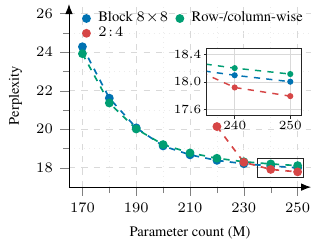}
  \caption{Perplexity vs. parameter count on Llama-350M.}
  \label{fig:pattern_ppl_params_350m}
\end{wrapfigure}

\textbf{Comparison of sparsity patterns.}
Figure~\ref{fig:pattern_ppl_params_350m} compares perplexity against parameter count for block, \mbox{$2\!:\!4$}, and row-/column-wise sparsity on Llama-350M.
Although $2\!:\!4$ sparsity achieves lower perplexity at the largest budgets, its fixed sparse capacity restricts elastic compression, with perplexity rising sharply as the parameter count decreases.
Block and row-/column-wise sparsity instead maintain competitive perplexity over a wider budget range.

\Needspace{13\baselineskip}

Heatmaps of the sparse components learned with S-SALAAD reveal pronounced row- or column-aligned patterns in attention projections (Appendices~\ref{app:training_patterns} and~\ref{app:paired_sparsity}).
This organization helps explain why the additional pruning freedom of blocks does not yield a clear advantage over row-/column-wise sparsity in the perplexity--parameter curves.
With comparable quality trade-offs, the efficient mapping of block computations onto current hardware motivates our focus on block sparsity and the development of dedicated CPU and GPU kernels.
This choice reflects current hardware and deployment requirements; other patterns may become preferable as these evolve, and S-SALAAD's unified formulation accommodates such alternatives.

\textbf{Block granularity and model scale.}
\begin{wrapfigure}{R}{0.60\linewidth}
  \centering
  \captionsetup{type=figure}
  \captionsetup[subfigure]{skip=2pt,font=small}
  \newlength{\blockscalingheight}
  \setlength{\blockscalingheight}{0.51\linewidth}
  \begin{subfigure}[b]{0.55\linewidth}
    \centering
    \includegraphics[height=\blockscalingheight]{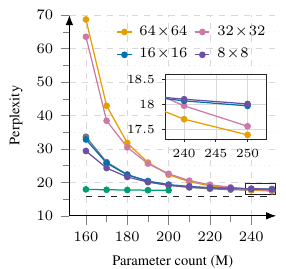}
    \caption{Llama-350M.}
    \label{fig:pvp_350m}
  \end{subfigure}\hfill%
  \begin{subfigure}[b]{0.43\linewidth}
    \centering
    \includegraphics[height=\blockscalingheight]{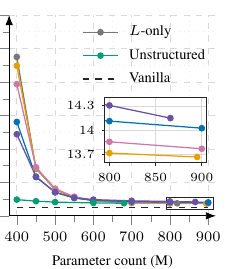}
    \caption{Llama-1B.}
    \label{fig:pvp_1b}
  \end{subfigure}
  \caption{Perplexity versus post-compression parameter count for different block sizes on Llama.}
  \label{fig:block_scaling}
\end{wrapfigure}
Figure~\ref{fig:block_scaling} compares the perplexity--parameter trade-offs of different block sizes on Llama-350M and Llama-1B, including $L$-only (see Appendix~\ref{sec:ablation}).
Within each model scale, the variants share the same base training configuration, with their settings for structure induction detailed in Appendix~\ref{app:pretraining}.
Retaining $S$ provides representational flexibility beyond the low-rank component, helping preserve model quality as the parameter budget tightens.
Finer blocks impose weaker structural constraints and allow more flexible allocation of sparse capacity during compression.
This benefit is evident for the $8\!\times\!8$ and $16\!\times\!16$ variants on both Llama-350M and Llama-1B at tight budgets (Figures~\ref{fig:pvp_350m} and~\ref{fig:pvp_1b}).
For example, at 500M parameters on Llama-1B, their perplexities are 17.07 and 17.03, respectively, compared with 17.86 for $L$-only.

\emph{This representational advantage, however, does not automatically translate into lower inference latency.}
Dense matrix kernels efficiently support the low-rank path, while fine-grained sparse inference introduces additional overhead.
At the 500M operating point on Llama-1B, the $16\!\times\!16$ variant achieves lower perplexity but incurs 7.9\% higher latency than $L$-only at approximately the same peak memory (Figures~\ref{fig:pvp_1b} and~\ref{fig:e2e_gpu}).
Similar overhead is observed on the RTX~2080~Ti and CPU (Figures~\ref{fig:e2e_2080ti_1b} and~\ref{fig:e2e_cpu_1b}).
Increasing block size trades some compression flexibility for more efficient inference.
With $64\!\times\!64$ blocks, the perplexity--memory--latency trade-offs closely track those of $L$-only on both GPUs (Figures~\ref{fig:e2e_gpu} and~\ref{fig:e2e_2080ti_1b}), while improving on them at CPU operating points (Figure~\ref{fig:e2e_cpu_1b}).
These results demonstrate that an appropriate block granularity enables the sparse component to improve deployment trade-offs beyond the $L$-only special case.
We note that the block-size benchmark in Appendix~\ref{app:kernel_speedup} and additional comparisons in Appendix~\ref{app:block_granularity_pareto} further characterize this trade-off.
The granularity that approximately preserves the $L$-only perplexity--parameter trade-off increases from $16\!\times\!16$ on Llama-350M to $64\!\times\!64$ on Llama-1B (Figure~\ref{fig:block_scaling}).
This trend suggests that larger models can accommodate coarser, more efficient blocks while retaining elastic compressibility, increasing the deployment benefits of structured $S$ representations at larger scales.

\section{Conclusion}
\label{sec:conclusion}

We presented S-SALAAD, a plug-and-play framework that learns structured SLR representations during training for elastic edge deployment.
Zero-shot budget-aware compression and dedicated block-sparse CPU and GPU kernels translate this learned compressibility into memory savings and efficient inference across budgets from a single checkpoint.
Across Llama pretraining and Qwen distillation, S-SALAAD advances the empirical perplexity-memory-latency Pareto frontier relative to the evaluated baselines.
Our analysis further links tolerable block granularity to model scale, providing empirical guidance for extending the framework to larger models.

\clearpage

\section*{Use of Large Language Models}
\label{app:llm}

In this work, we used generative AI tools to provide feedback on research methodology and experiments, implement methods, translate text, and clean and reformat datasets.
Additionally, we used generative AI tools to create and edit plotting code and scientific figures, and to edit the manuscript for readability.
We did not use generative AI tools to design the research methodology or experiments, develop theoretical models or conceptual frameworks, formulate mathematical claims, provide critical ingredients for or assist in writing mathematical proofs, propose or refine hypotheses, or interpret results.
Synthetic-data generation and qualitative or thematic data analysis were not applicable to this work.
All AI-assisted work was reviewed by the authors, and the implementations, processed data, plotting code, and resulting figures were verified by two authors against the intended methods, source data, and experimental records.
We take responsibility for the final content of this work, including text, claims, code, and artifacts produced with the aid of generative AI.

\section*{Reproducibility Statement}
\label{app:reproducibility}

The learning objective and optimization updates are specified in Sections~\ref{sec:problem} and~\ref{sec:algorithm}, and Algorithms~\ref{alg:ssalaad_pretrain} and~\ref{alg:iterative_svd} describe training and I-SVD, respectively.
Section~\ref{sec:implementation} and Appendix~\ref{app:end_to_end_details} document compression, inference, and supporting implementation details.
Appendix~\ref{app:setup} provides the datasets and tokenization, model configurations, training hyperparameters, parameter counts, baseline implementations, and evaluation protocols, including hardware, precision, input shapes, timing, and memory measurement.
\ifshowresources
Model checkpoints, teacher logits, and the codebase are available on the \href{https://egretwalker.github.io/S-SALAAD/}{project website}.
\else
Model checkpoints, teacher logits, and the codebase will be released.
\fi

\bibliography{iclr2027_conference}
\bibliographystyle{iclr2027_conference}

\clearpage
\appendix

\section{Motivation for Structured Sparsity During Pretraining}
\label{app:deployment_barriers}

\subsection{Memory Savings Do Not Guarantee Lower Inference Latency}
\label{app:compression_latency_gap}

\begin{figure}[H]
  \centering
  \includegraphics[width=0.48\linewidth]{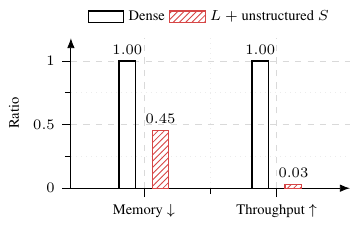}
  \caption{Static weight memory and throughput of dense inference and the unstructured SLR representation used by SALAAD. Both metrics are normalized to their respective dense baselines. The benchmark uses a synthetic Llama-1B MLP down projection with a \qty{15}{\percent} rank ratio and \qty{5}{\percent} nonzero density.}
  \label{fig:salaad_overhead}
\end{figure}

We compare dense inference with the unstructured SLR representation used by SALAAD on a synthetic Llama-1B MLP down projection of size $2048\!\times\!5632$.
The benchmark uses a \qty{15}{\percent} rank ratio and \qty{5}{\percent} nonzero density, with the sparse component stored as a PyTorch \texttt{torch.sparse\_csr\_tensor}.
Both representations are profiled on a Quadro RTX~6000 with batch size~128 and sequence length~256.
As shown in \figref{fig:salaad_overhead}, static weight memory falls from \qty{23.07}{\mega\byte} to \qty{10.49}{\mega\byte}, a \qty{54.5}{\percent} reduction, while throughput falls to \qty{2.68}{\percent} of the dense reference.
For this fixed workload, the throughput ratio implies a projection inference latency $37.3$ times that of the dense baseline.
Unstructured sparse components incur indexing and irregular memory-access overheads that can offset the computational savings from sparsity~\citep{mishra2021accelerating}.
This example illustrates why reduced weight storage alone does not ensure efficient inference.

\subsection{Quality Degradation from Imposing Block Structure After Pretraining}
\label{app:salaad_as_blocks}

We test post-hoc block compression by applying energy-guided alternating refinement (EGAR; Section~\ref{sec:implementation}) to the same SALAAD Llama-1B checkpoint pretrained with unstructured sparsity, under different block constraints and parameter budgets.
Table~\ref{tab:salaad_as_blocks} reveals two trends.
First, at a fixed budget, perplexity rises as the imposed blocks become coarser.
Second, with coarser blocks, perplexity increases more steeply as the parameter budget decreases.

These trends are consistent with the reduced pruning flexibility of coarser blocks.
Grouping more weights into a single pruning unit restricts which weights can be retained at a given budget, limiting the compressed model's representational capacity.
A checkpoint pretrained with unstructured sparsity has not adapted to these constraints, making post-hoc pruning more damaging as the budget tightens.
These results motivate inducing the target structure during pretraining, allowing the sparse and low-rank components to adapt to it under the task loss.

\begin{table}[H]
\centering
\small
\caption{Post-compression perplexity of the same unstructured SALAAD Llama-1B checkpoint when block structure of size $p\!\times\!p$ is imposed during compression, at three parameter budgets $P_{\mathrm{tgt}}$. The first row is the element-wise reference.}
\label{tab:salaad_as_blocks}
\begin{tabular}{@{}l r r r@{}}
\toprule
Block size& $P_{\mathrm{tgt}}\!=\!500$M & $P_{\mathrm{tgt}}\!=\!600$M & $P_{\mathrm{tgt}}\!=\!700$M \\
\midrule
$1\!\times\!1$ & 14.11 & 13.86 & 13.78 \\
$2\!\times\!2$          & 17.29 & 14.93 & 14.19 \\
$4\!\times\!4$          & 87.44 & 28.89 & 18.28 \\
$8\!\times\!8$          & 456.52 & 116.33 & 43.58 \\
$16\!\times\!16$        & 1497.68 & 351.39 & 98.83 \\
\bottomrule
\end{tabular}
\end{table}

\section{$L$-Only S-SALAAD}
\label{sec:ablation}

The $L$-only variant is the $S\equiv0$ special case of S-SALAAD.
It retains ADMM's coupled updates of $X$ and $L$ under the constraint $X\!=\!L$, with the proximal update~\eqref{eq:lstep} explicitly removing small singular values from $L$.
This structure-learning procedure goes beyond directly adding a low-rank regularizer to the task loss.

\begin{figure}[h]
  \centering
  \includegraphics[width=0.48\linewidth]{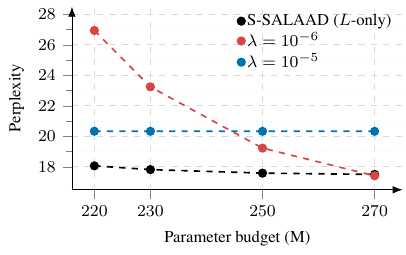}
  \caption{Post-compression perplexity on Llama-350M.}
  \label{fig:reg_ablation}
\end{figure}

Figure~\ref{fig:reg_ablation} compares this variant with direct subgradient optimization of $\ell(X)\!+\!\lambda|X|_*$.
With $\lambda\!=\!10^{-6}$, perplexity is comparable to that of $L$-only S-SALAAD at the largest budget but rises sharply as the budget decreases.
This behavior is consistent with low-rank structure that is too weak to support aggressive compression.
Increasing $\lambda$ to $10^{-5}$ strengthens rank regularization and yields nearly constant perplexity across the tested budgets, supporting elastic compression at the cost of higher perplexity throughout.
$L$-only S-SALAAD maintains low perplexity over the same budget range, providing a better trade-off between model quality and elastic compressibility than these direct regularization baselines.

\section{Additional Training and Compression Details}
\label{app:training_details}
\label{app:end_to_end_details}

\subsection{Adaptive I-Controller}
\label{app:i_controller}
\label{app:allocation_statistics}

S-SALAAD follows the integral controller (I-controller) update rule of SALAAD~\citep{ma2026salaad}, which we briefly restate here for completeness.
We first define the effective-rank ratio and group sparsity used by the controller.

For a low-rank component $L\!\in\!\R^{n\times m}$, let $r_0\!\doteq\!\min(n,m)$ and let $\sigma_1\!\geq\!\cdots\!\geq\!\sigma_{r_0}\!\geq\!0$ denote its singular values.
For $L\neq0$ and an energy-coverage level $\gamma\!\in\!(0,1]$, we define
\begin{equation}
  i_\gamma
  \doteq
  \min\left\{
    i\in\{1,\ldots,r_0\}:
    \frac{\sum_{j=1}^{i}\sigma_j^2}{\sum_{j=1}^{r_0}\sigma_j^2}
    \geq \gamma
  \right\},
  \quad
  \Gamma_L \doteq \frac{i_\gamma}{r_0}.
  \label{eq:effective_rank_ratio}
\end{equation}
When $L\!=\!0$, we set $i_\gamma\!=\!\Gamma_L\!=\!0$.
Thus, $\Gamma_L$ is the fraction of available singular directions required to retain a fraction $\gamma$ of the energy in $L$.
For a sparse component $S$ partitioned into the $D$ non-overlapping groups $\{S^d\}_{d=1}^{D}$ of $\Omega$, its sparsity is the fraction of groups that are zero,
\begin{equation}
  \Upsilon_S \doteq \frac{\sum_{d=1}^{D}\mathds{1}\big\{|S^{d}|_\mathrm{F}=0\big\}}{D}.
  \label{eq:sparse_density}
\end{equation}
Evaluating the statistic over groups rather than entries matches the granularity at which the $S$-update removes parameters; for $D\!=\!nm$ singleton groups, this reduces to the element-wise sparsity $1-\nicefrac{|S|_0}{nm}$.
Both statistics are used by the I-controller during training and by Stage~I of EGAR, which recomputes them after every thresholding step and uses $G\!\doteq\!\Gamma_L\!+\!(1-\Upsilon_S)$ to adapt the next threshold update to the current low-rank and sparse components.

After each ADMM round, the controller adjusts the regularization strengths $\alpha$ and $\beta$ so that the induced effective-rank ratio $\Gamma_L$ and group sparsity $\Upsilon_S$ track targets $(\widehat{\Gamma},\widehat{\Upsilon})$.
Given controller step sizes $(\Delta_\alpha,\Delta_\beta)$, the update is
\begin{equation}
  \alpha \leftarrow \alpha+\rho\left(\Gamma_L-\widehat{\Gamma}\right)\Delta_\alpha,
  \quad
  \beta \leftarrow \beta+\rho\left(\widehat{\Upsilon}-\Upsilon_S\right)\Delta_\beta .
  \label{eq:i_controller_app}
\end{equation}
When $\Gamma_L$ exceeds its target, increasing $\alpha$ raises the low-rank proximal threshold $\nicefrac{\alpha}{\rho}$; when $\Upsilon_S$ falls below its target, increasing $\beta$ raises $\nicefrac{\beta}{\rho}$ and induces more zero groups.
The corresponding penalty decreases in the opposite case.
The factor $\rho$ makes the increments of the effective thresholds independent of the ADMM penalty, since dividing \eqref{eq:i_controller_app} by $\rho$ yields the increments $\left(\Gamma_L\!-\!\widehat{\Gamma}\right)\Delta_\alpha$ and $\left(\widehat{\Upsilon}\!-\!\Upsilon_S\right)\Delta_\beta$.
Under the $N\!:\!M$ regularizer, where $\Omega$ carries group-specific coefficients $\{\beta_d\}_{d=1}^{D}$, the $\beta$-update in \eqref{eq:i_controller_app} is applied per group, with $\Upsilon_S$ replaced by the element-wise sparsity $\Upsilon_{S^{d}}\!=\!1-\nicefrac{|S^d|_0}{M}$ within the $d$-th group and target $\widehat{\Upsilon}\!=\!1-\nicefrac{N}{M}$; the $\alpha$-update is unchanged.
The selection of $\rho$ and the controller step sizes is described in Appendix~\ref{app:hparams}.

\subsection{Fidelity of Iterative SVD}
\label{app:iterative_svd_results}

\begin{figure}[H]
  \centering
  \includegraphics[width=0.48\linewidth]{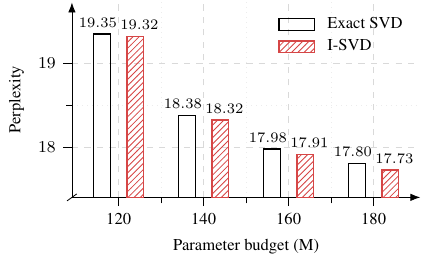}
  \caption{Post-compression perplexity for exact and iterative SVD on Llama-350M at matched budgets.}
  \label{fig:svdt_ppl}
\end{figure}

\begin{figure}[H]
  \centering
  \begin{subfigure}[b]{0.32\linewidth}
    \centering
    \includegraphics[width=\linewidth]{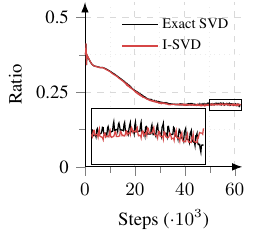}
    \caption{Layer 4.}
    \label{fig:svdt_curves_l4}
  \end{subfigure}\hfill
  \begin{subfigure}[b]{0.32\linewidth}
    \centering
    \includegraphics[width=\linewidth]{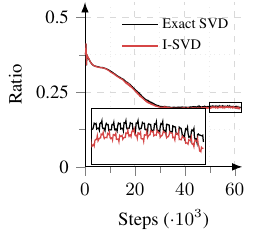}
    \caption{Layer 8.}
    \label{fig:svdt_curves}
  \end{subfigure}\hfill
  \begin{subfigure}[b]{0.32\linewidth}
    \centering
    \includegraphics[width=\linewidth]{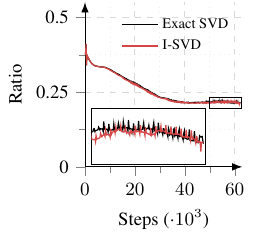}
    \caption{Layer 14.}
    \label{fig:svdt_curves_l14}
  \end{subfigure}
  \caption{$\Gamma_L$ trajectories for exact and iterative SVD on the down projections of three representative Llama-350M layers. Each inset magnifies the converged regime, where the two curves are otherwise indistinguishable.}
  \label{fig:warm_started_svd}
\end{figure}

Figures~\ref{fig:svdt_ppl} and~\ref{fig:warm_started_svd} compare I-SVD with exact SVD beyond the training-time results reported in the main text.
I-SVD preserves both the post-compression validation perplexity and the effective-rank trajectories of exact SVD, confirming that the approximation does not degrade the learned decomposition.

\subsection{Compression by Energy-Guided Alternating Refinement}
\label{app:energy_guided_alternating_refinement}

\begin{figure}[t]
  \centering
  \newlength{\egarcomparisonheight}
  \setlength{\egarcomparisonheight}{0.32\linewidth}
  \begin{minipage}[t]{0.49\linewidth}
    \vspace{0pt}
    \centering
    \includegraphics[height=\egarcomparisonheight]{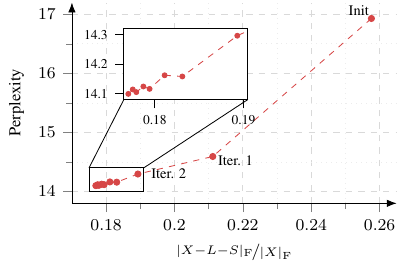}
    \caption{Alternating refinement trajectory on Llama-1B pretrained with unstructured SALAAD at a 500M parameter budget.}
    \label{fig:energy_guided_alternating_refinement}
  \end{minipage}\hfill
  \begin{minipage}[t]{0.49\linewidth}
    \vspace{0pt}
    \centering
    \includegraphics[height=\egarcomparisonheight]{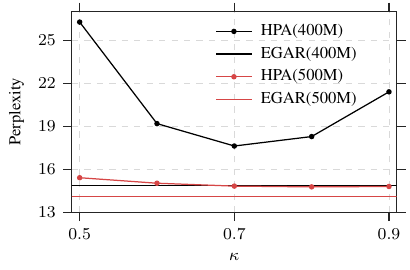}
    \caption{HPA versus EGAR on Llama-1B pretrained with unstructured SALAAD at 400M and 500M parameter budgets.}
    \label{fig:hpa_vs_egar}
  \end{minipage}
\end{figure}

The following EGAR ablations use Llama-1B pretrained with unstructured SALAAD~\citep{ma2026salaad}.
Figure~\ref{fig:energy_guided_alternating_refinement} traces Stage~II on Llama-1B at a 500M parameter budget.
Starting from the allocation produced by Stage~I, alternating refinement monotonically reduces the normalized reconstruction error, while validation perplexity falls sharply and then stabilizes with minor fluctuations.
Most of the improvement occurs within the first iterations, and the procedure converges in fewer than 10 iterations, taking only a few minutes in total.
We compare EGAR with \emph{homomorphic parameter allocation} (HPA), the greedy pruning strategy used in SALAAD~\citep{ma2026salaad}.
HPA assumes homomorphic shrinkage across linear projections and performs single-pass greedy pruning after splitting $P_{\mathrm{tgt}}$ between sparse and low-rank components with a rigid ratio $\kappa$, thereby ignoring the heterogeneity across linear projections and potentially degrading performance.
Both methods compress the same Llama-1B checkpoint pretrained with unstructured SALAAD, as shown in \figref{fig:hpa_vs_egar}.
At the 400M and 500M parameter budgets, EGAR achieves lower perplexity than HPA for all evaluated values of $\kappa$ between $0.5$ and $0.9$.

\subsection{Kernel Micro-Benchmark}
\label{app:kernel_speedup}

\begin{figure}[H]
  \centering
  \captionsetup[subfigure]{skip=2pt,font=small,justification=centering}
  \newlength{\kernelpanelheight}
  \setlength{\kernelpanelheight}{0.178\linewidth}
  \begin{subfigure}[t]{0.24\linewidth}
    \vspace{0pt}
    \centering
    \includegraphics[height=\kernelpanelheight]{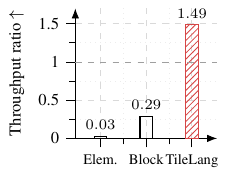}
    \caption{GPU throughput.}
    \label{fig:kernel_throughput_gpu}
  \end{subfigure}\hfill
  \begin{subfigure}[t]{0.24\linewidth}
    \vspace{0pt}
    \centering
    \includegraphics[height=\kernelpanelheight]{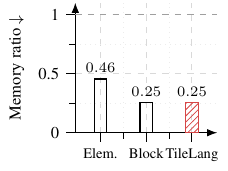}
    \caption{GPU static weight memory.}
    \label{fig:kernel_memory_gpu}
  \end{subfigure}\hfill
  \begin{subfigure}[t]{0.24\linewidth}
    \vspace{0pt}
    \centering
    \includegraphics[height=\kernelpanelheight]{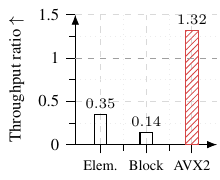}
    \caption{CPU throughput.}
    \label{fig:kernel_throughput_cpu}
  \end{subfigure}\hfill
  \begin{subfigure}[t]{0.24\linewidth}
    \vspace{0pt}
    \centering
    \includegraphics[height=\kernelpanelheight]{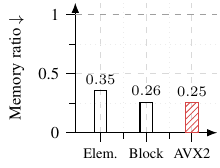}
    \caption{CPU static weight memory.}
    \label{fig:kernel_memory_cpu}
  \end{subfigure}
  \caption{Kernel efficiency on a randomly sparsified Llama-1B MLP down projection at \qty{15}{\percent} rank ratio and \qty{5}{\percent} density, normalized to dense execution. Elem. and Block denote PyTorch's \texttt{torch.sparse\_csr\_tensor} and \texttt{torch.sparse\_bsr\_tensor} backends, respectively. In the block-sparse setting both backends run at $16\!\times\!16$ on the GPU and $8\!\times\!8$ on the CPU, the smallest block size each platform supports.}
  \label{fig:kernel_speedup}
\end{figure}

On a randomly sparsified Llama-1B linear projection (\figref{fig:kernel_speedup}), unstructured sparsity incurs prohibitive compute overhead and moderate memory overhead.
Stock PyTorch block-sparse SpMM\footnote{PyTorch~2.13.0 with the CUDA~12.9 build.} is in beta version. In practice we find it uncompetitive with the dense baseline on either platform. Our custom TileLang (GPU) and AVX2 (CPU) kernels realise the structural savings and deliver measured speedups.

\begin{figure}[H]
  \centering
  \includegraphics[width=0.55\linewidth]{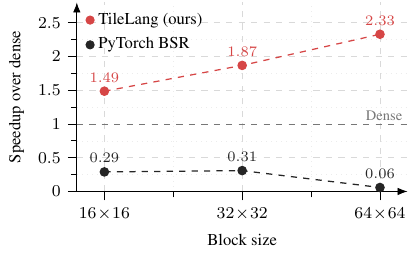}
  \caption{Block-size sweep on random SLR weights for a $2048\!\times\!5632$ Llama-1B down projection, measured on a Quadro~RTX~6000 in FP16 with batch size~128 and sequence length~256. Throughput is normalized to dense.}
  \label{fig:block_size_kernel_gpu}
\end{figure}

\textbf{Block size and inference speed.}
Figure~\ref{fig:block_size_kernel_gpu} compares $16\!\times\!16$, $32\!\times\!32$, and $64\!\times\!64$ blocks on randomly generated SLR projections with rank~307 and a target nonzero density of \qty{5}{\percent}; block masks are sampled independently at each size.
With our TileLang kernel, throughput relative to dense increases from $1.49\times$ to $1.87\times$ and $2.33\times$, respectively, giving $64\!\times\!64$ blocks $1.57\times$ the throughput of $16\!\times\!16$ blocks in this setting.
PyTorch BSR remains slower than dense at all three sizes, highlighting the role of the kernel in realizing the benefits of coarser blocks.

\section{Implementation Details}
\label{app:setup}

\subsection{Relevance to Edge Deployment}
\label{app:edge_deployment_scope}

This work focuses on reducing the memory footprint and inference latency of large models for edge deployment.
Our evaluation uses single-GPU inference on a Quadro~RTX~6000 with 24~GB of GPU memory and an RTX~2080~Ti with 11~GB of GPU memory, together with CPU-only inference on 4$\times$ AMD~EPYC~7662 CPUs with 32~GB of RAM.
Edge computing encompasses computation near data sources, including local compute nodes~\citep{shi2016edge}.
We use these workstation/server platforms as proxies for edge inference in terms of memory capacity and available acceleration, the resource constraints targeted by S-SALAAD.
Energy consumption and thermal management are outside the scope of this study.

\subsection{Datasets and Model Configurations}
\label{app:datasets}
\label{app:arch}
\label{app:arch-llama}
\label{app:arch-qwen}

\textbf{Datasets and tokenization.}
For Llama 2 pretraining from scratch we use the C4 corpus \citep{raffel2020c4} tokenised with \texttt{t5-base}.
For the cross-architecture distillation of Qwen3-1.7B we use FineWeb-Edu~\citep{penedo2024fineweb}, tokenised with the teacher's Nemotron tokenizer so that student and teacher share a token space.

\textbf{Llama 2 pretraining sizes.}
Table~\ref{tab:model-arch} lists the Llama 2 architectures we pretrain.
Llama-1B supports the main block-sparse experiments, while Llama-350M is used for comparisons among sparsity patterns.
We compare block granularities across both model sizes.

\begin{table}[H]
  \centering
  \small
  \caption{Model architectures used for Llama 2 pretraining and Qwen3 distillation.}
  \label{tab:model-arch}
  \begin{tabularx}{\linewidth}{@{}l*{3}{>{\centering\arraybackslash}X}@{}}
  \toprule
  Setting & \textbf{Llama-350M} & \textbf{Llama-1B} & \textbf{Qwen3-1.7B} \\
  \midrule
  Hidden size                & 1024 & 2048 & 2048 \\
  Intermediate (FFN) size    & 2816 & 5632 & 6144 \\
  Layers                     & 24 & 24 & 28 \\
  Attention heads (Q / KV)   & 16 / 16 (MHA) & 32 / 32 (MHA) & 16 / 8 (GQA) \\
  Head dimension             & 64 & 64 & 128 \\
  Vocabulary size            & 32,128 & 32,128 & 131,072 \\
  Activation                 & \multicolumn{3}{c@{}}{SwiGLU} \\
  Normalisation              & \multicolumn{3}{c@{}}{RMSNorm ($\varepsilon = 10^{-6}$), pre-norm} \\
  QK-normalisation           & no & no & yes \\
  Positional encoding        & \multicolumn{2}{c}{RoPE, $\theta = 10^{4}$} & RoPE, $\theta = 10^{6}$ \\
  Tied I/O embeddings        & no & no & yes \\
  \bottomrule
  \end{tabularx}
  \end{table}
\textbf{Qwen3-1.7B distillation student.}
We adopt the standard Qwen3-1.7B backbone \citep{yang2025qwen3} with a single modification (see Table~\ref{tab:model-arch}): the input/output embedding vocabulary is swapped from Qwen3's native one to the teacher's Nemotron tokenizer, so that student and teacher share a token space and the top-$K$ KL distillation loss is well-defined without a re-tokenisation step.

%

\subsection{Training Configurations}
\label{app:hparams}
\label{app:pretraining}
\label{app:distill}

\textbf{Pretraining from scratch.}
Table~\ref{tab:pretrain-hparams} summarizes the shared and model-specific settings for Llama 2 pretraining.

\textbf{Cross-architecture distillation.}
The student uses the teacher's tokenizer and is trained with the distillation loss~\citep{peng2025ptdistill}
\begin{equation*}
  \mathcal{L}_{\mathrm{distill}}
  = (1-\alpha_{\mathrm{CE}})\,
    \operatorname{KL}_{\mathrm{top}\text{-}K}(p_{\mathrm{T}}\,\|\,p_{\mathrm{S}})
    + \alpha_{\mathrm{CE}}\,\operatorname{CE}(y,p_{\mathrm{S}}),
\end{equation*}
where $p_{\mathrm{T}}$ and $p_{\mathrm{S}}$ denote the teacher and student next-token distributions, respectively, and $y$ is the ground-truth next token.
Table~\ref{tab:distill-hparams} lists the settings used for the cross-architecture distillation.

\begin{table}[H]
\centering
\small
\setlength{\tabcolsep}{4pt}
\begin{minipage}[t]{0.49\linewidth}
\vspace{0pt}
\caption{Llama 2 pretraining settings.}
\label{tab:pretrain-hparams}
\centering
\begin{tabular*}{\linewidth}{@{\extracolsep{\fill}}lcc@{}}
\toprule
Setting & 350M & 1B \\
\midrule
Optimizer            & \multicolumn{2}{c@{}}{Adam} \\
$(\beta_1, \beta_2)$ & \multicolumn{2}{c@{}}{$(0.9, 0.95)$} \\
$\varepsilon$        & \multicolumn{2}{c@{}}{$10^{-8}$} \\
LR schedule          & \multicolumn{2}{c@{}}{\begin{tabular}[t]{@{}l@{}}cosine, \qty{10}{\percent} warmup,\\$\min$-LR ratio $0.1$\end{tabular}} \\
Batch size           & \multicolumn{2}{c@{}}{512 sequences} \\
Sequence length      & \multicolumn{2}{c@{}}{256} \\
Tokenizer            & \multicolumn{2}{c@{}}{\texttt{t5-base}} \\
Precision            & \multicolumn{2}{c@{}}{bf16 mixed} \\
\midrule
Peak LR              & $10^{-3}$ & $5\!\times\!10^{-4}$ \\
Total steps          & 62,250 & 128,000 \\
Tokens (B)           & 8.16 & 16.78 \\
\bottomrule
\end{tabular*}
\end{minipage}\hfill
\begin{minipage}[t]{0.49\linewidth}
\vspace{0pt}
\caption{Cross-architecture distillation.}
\label{tab:distill-hparams}
\centering
\begin{tabularx}{\linewidth}{@{}l>{\raggedright\arraybackslash}X@{}}
\toprule
Setting & Value \\
\midrule
Teacher                 & Nemotron-3-Super-120B (NVFP4) \\
Corpus                  & FineWeb-Edu \\
Tokenizer               & Nemotron (teacher's) \\
$\alpha_\text{CE}$      & 0.1 \\
Top-$K$                 & 10 \\
Optimizer               & AdamW \\
$(\beta_1,\beta_2)$     & $(0.9, 0.95)$ \\
Peak LR                 & $4\!\times\!10^{-4}$ \\
Schedule                & cosine, \qty{10}{\percent} warmup, $\min$-LR ratio $0.1$ \\
Total steps             & 180,000\\
Batch size             & 512 sequences \\
Sequence length         & 512 \\
\bottomrule
\end{tabularx}
\end{minipage}
\end{table}

\textbf{Hyperparameter settings.}
We use a target effective-rank ratio $\widehat{\Gamma}\!=\!0.15$ and target sparsity $\widehat{\Upsilon}\!=\!0.95$.
The $L$-only special case retains the same rank target but omits the sparse target and $\beta$-update, while $2\!:\!4$ sparsity uses $\widehat{\Gamma}\!=\!0.03$ and $\widehat{\Upsilon}\!=\!0.5$.
We fix the energy-coverage level $\gamma\!=\!0.999$ throughout.
For the row-/column-wise variant, we sparsify rows in the query, key, value, gate, and up projections and columns in the output and down projections.

We set $\rho$ according to the empirical scaling rule proposed in SALAAD~\citep{ma2026salaad}, 
\begin{equation}
  \label{eq:rho_scaling}
  \rho \propto \frac{1}{E\sqrt{nm}},
\end{equation}
where $E$ is the number of selected linear projections and $n\!\times\!m$ is the shape of the current projection's weight matrix.
The unstructured baseline uses the $\Delta_\beta$ setting from SALAAD.
To induce structured sparsity, we increase $\Delta_\beta$ and correspondingly decrease $\Delta_\alpha$.
Within each sparsity pattern, we make only minor adjustments to the controller step sizes $\Delta_\alpha$ and $\Delta_\beta$, whose values are reported in Table~\ref{tab:variant-hparams}.

Our experiments indicate that, within a given sparsity variant, $\Delta_\alpha$ and $\Delta_\beta$ can generally be reused across model scales, with $\rho$ alone adjusted according to the scaling rule above.
This observation is consistent with SALAAD's hyperparameter-scaling strategy~\citep{ma2026salaad} and provides an empirical basis for extending S-SALAAD to larger models.

\begin{table}[H]
\centering
\small
\caption{Controller step sizes across model scales and sparsity variants.}
\label{tab:variant-hparams}
\begin{tabular}{@{}l cc cc cc@{}}
\toprule
\multirow{2}{*}{Variant} & \multicolumn{2}{c}{\textbf{Llama-350M}} & \multicolumn{2}{c}{\textbf{Llama-1B}} & \multicolumn{2}{c}{\textbf{Qwen3-1.7B}} \\
\cmidrule(lr){2-3}\cmidrule(lr){4-5}\cmidrule(l){6-7}
 & $\Delta_\alpha$ & $\Delta_\beta$ & $\Delta_\alpha$ & $\Delta_\beta$ & $\Delta_\alpha$ & $\Delta_\beta$ \\
\midrule
Unstructured SALAAD                 & 0.20 & 0.001 & 0.20  & 0.001 & ---  & ---  \\
Block-$8\!\times\!8$                & 0.04 & 0.01  & 0.045 & 0.01  & ---  & ---  \\
Block-$16\!\times\!16$              & 0.04 & 0.01  & 0.04  & 0.01  & ---  & ---  \\
Block-$32\!\times\!32$              & 0.04 & 0.01  & 0.04  & 0.01  & ---  & ---  \\
Block-$64\!\times\!64$              & 0.02 & 0.02  & 0.02  & 0.02  & 0.02 & 0.02 \\
$L$-only ($S \equiv 0$)             & 0.03 & ---   & 0.03  & ---   & ---  & ---  \\
$2\!:\!4$ semi-structured           & 0.10 & 0.025 & ---   & ---   & ---  & ---  \\
Row-/column-wise                    & 0.04 & 0.04  & ---   & ---   & ---  & ---  \\
\bottomrule
\end{tabular}
\end{table}

\subsection{Evaluation and Baselines}
\label{app:evaluation}
\label{app:baselines}

\textbf{Quality evaluation.}
Except for the separately sourced LORO reference described below, perplexity is measured on held-out tokens of the training corpus---C4 for Llama and FineWeb-Edu for Qwen3-1.7B---under the corresponding evaluation protocol.
Perplexities are compared within each model family, not across the two corpora.

We evaluate zero-shot downstream accuracy on ARC-Easy, OpenBookQA, WinoGrande, HellaSwag, PIQA, and SciQ using the LM Evaluation Harness~\citep{gao2024lmeval}.

\textbf{Baseline implementations and profiling.}
Vanilla training and SALAAD~\citep{ma2026salaad} provide training references, while the $L$-only special case ($S\equiv0$) provides a reference for block granularity.
EGAR produces variants at multiple parameter budgets from each S-SALAAD checkpoint without calibration data.
All compressed models are evaluated without additional fine-tuning for fair comparison.

For Wanda and SparseGPT, we use the measured latency and throughput of the corresponding dense models, as the evaluated hardware lacks native acceleration for $2\!:\!4$ sparsity.
This illustrates that the latency benefits of a sparsity pattern are constrained by support from the target hardware.

We use the SVD-LLM~V2 implementation~\citep{wang2025svdllm2} for the SVD-LLM baseline.
For LORO, we use its published Llama-1B perplexity of 15.19 at rank~512 after training on 13.1B C4 tokens~\citep{mo2025loro}.
We profile its deployment form locally by replacing the seven projections in each layer of our Llama-1B backbone with pairs of rank-512 linear factors, retaining dense embeddings and the output head.
This backbone uses FFN width~5632, compared with 5461 in the published LORO model, so the combined quality--cost point serves as an external reference.

SliceGPT and SVD-LLM inference costs are likewise profiled with synthetic models of the corresponding compressed dimensions; for Llama-1B, the same cost measurements are used for their standard-pretrained and S-SALAAD-pretrained variants.
The synthetic models use the same precision, input shapes, and forward-timing path as the checkpoint-based models, with the corresponding dense or factorized projection operators.
For RTX~2080~Ti profiling of $L$-only and the $16\!\times\!16$ and $32\!\times\!32$ variants, we retain the compressed components and randomly initialize the remaining parameters, while evaluating perplexity on the corresponding trained checkpoints.

\subsection{Profiling Environment and Protocol}
\label{app:compute}

The main GPU results use a Quadro~RTX~6000, with additional results on an RTX~2080~Ti. Table~\ref{tab:profiling-env} details the environment for the Quadro~RTX~6000 and CPU measurements; CPU runs use 4$\times$ AMD EPYC 7662 CPUs.

\begin{table}[H]
\centering
\small
\caption{Profiling environment for the Quadro~RTX~6000 and CPU measurements.}
\label{tab:profiling-env}
\begin{tabular}{@{}l l@{}}
\toprule
Component & Value \\
\midrule
GPU               & NVIDIA Quadro RTX~6000 (Turing, TU102) \\
GPU memory        & 24~GB GDDR6 \\
GPU driver / CUDA & NVIDIA driver 550.x / CUDA 12.9 \\
CPU               & 4$\times$ AMD EPYC 7662 CPUs \\
RAM               & 32~GB \\
PyTorch           & 2.13.0 (cu129 build) \\
TileLang          & 0.1.12 \\
\bottomrule
\end{tabular}
\end{table}

\textbf{Inference profiling protocol.}
Table~\ref{tab:profiling-protocol} specifies the fixed input shapes and repetition counts used for the Pareto-frontier measurements.
Each run uses 10 untimed warmup passes followed by 20 timed passes, except for Qwen on CPU, which uses 3 warmup passes and 6 timed passes.
GPU profiling uses FP16 on both GPU models, while CPU profiling uses FP32 on the 4$\times$ AMD EPYC 7662 CPUs.
Models run in evaluation mode with gradients and KV caching disabled.
GPU profiling enables \texttt{torch.compile} in \texttt{reduce-overhead} mode; CPU profiling uses eager inference.

After untimed warmup, we measure each forward pass with a wall-clock timer, synchronizing CUDA immediately before and after the timed region on GPUs, and report the median latency.
Throughput is the number of input tokens per batch divided by this median latency.
We report results only after latency and throughput have stabilized; repeated profiling runs show negligible variation.
Data loading, input transfer, compilation, and loss evaluation are excluded from timing.

The timed forward computes logits only for the final two positions of each sequence; the perplexities plotted in the figures come from separate held-out evaluations, rather than the auxiliary last-token estimates recorded during profiling.
These measurements characterize prefill computation and do not include autoregressive decoding or complete request latency.

\begin{table}[H]
\centering
\small
\setlength{\tabcolsep}{4pt}
\caption{Input shapes and repetition counts for inference profiling.}
\label{tab:profiling-protocol}
\begin{tabular}{@{}l l c c c@{}}
\toprule
Hardware & Model & Batch $\times$ length & Warmup & Timed \\
\midrule
\multirow{2}{*}{Quadro~RTX~6000} & Llama-1B   & $128\times256$ & 10 & 20 \\
                                & Qwen3-1.7B & $32\times512$  & 10 & 20 \\
\midrule
\multirow{2}{*}{RTX~2080~Ti}     & Llama-1B   & $128\times256$ & 10 & 20 \\
                                & Qwen3-1.7B & $32\times512$  & 10 & 20 \\
\midrule
\multirow{2}{*}{4$\times$ AMD~EPYC~7662 CPUs} & Llama-1B   & $8\times64$ & 10 & 20 \\
                                    & Qwen3-1.7B & $8\times64$ & 3  & 6 \\
\bottomrule
\end{tabular}
\end{table}

\textbf{Peak-memory measurement.}
Peak memory is measured separately with an eager prefill pass that retains logits only at the final sequence position, without evaluating the loss.
On GPUs, we reset the peak-memory counter and report peak GPU memory usage as the maximum allocated tensor memory, including resident model weights, buffers, and sparse indices; allocator-reserved memory is excluded.
On CPU, we report peak RAM usage as the stored parameter and buffer sizes plus the increase in the process's peak resident memory during the forward pass.
This measures model storage and forward-pass working memory without compilation allocations or a persistent KV cache.

\begin{figure}[h]
  \centering
  \newlength{\appendixlossheight}
  \setlength{\appendixlossheight}{0.34\linewidth}
  \begin{subfigure}[t]{0.32\linewidth}
    \centering
    \includegraphics[height=\appendixlossheight]{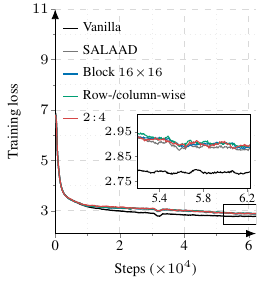}
    \caption{Llama-350M.}
    \label{fig:training_350m_loss}
  \end{subfigure}\hfill%
  \begin{subfigure}[t]{0.32\linewidth}
    \centering
    \includegraphics[height=\appendixlossheight]{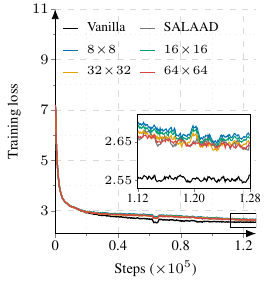}
    \caption{Llama-1B.}
    \label{fig:training_1b_loss}
  \end{subfigure}\hfill%
  \begin{subfigure}[t]{0.32\linewidth}
    \centering
    \includegraphics[height=\appendixlossheight]{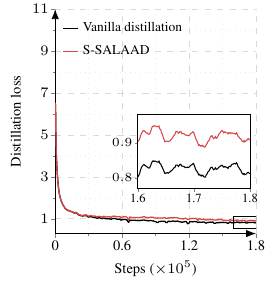}
    \caption{Qwen3-1.7B.}
    \label{fig:training_qwen_loss}
  \end{subfigure}
  \caption{Task-loss trajectories for (a) different sparsity patterns on Llama-350M, (b) different block sizes on Llama-1B, and (c) Qwen3-1.7B distillation. Curves use a moving average over approximately 1,000 training steps and end at the last available logged step.}
  \label{fig:training_losses}
\end{figure}

\section{Training Dynamics and Learned Sparse Structure}
\label{app:training_dynamics}

\subsection{Training Loss}
\label{app:training_loss}

Figure~\ref{fig:training_losses} compares loss trajectories across model sizes, sparsity patterns, block granularities, and training settings.
On Llama-350M, we examine block-$16\!\times\!16$, row-/column-wise, and $2\!:\!4$ sparsity under the settings in Appendix~\ref{app:pretraining}, with vanilla pretraining and unstructured SALAAD as references (Figure~\ref{fig:training_350m_loss}).
For Llama-1B, Figure~\ref{fig:training_1b_loss} compares block sizes $8\!\times\!8$, $16\!\times\!16$, $32\!\times\!32$, and $64\!\times\!64$ with the same training references.
The curves come from the runs corresponding to the released checkpoints, retaining their original training configurations and controller gains.
Figure~\ref{fig:training_qwen_loss} compares vanilla distillation with block-$64\!\times\!64$ S-SALAAD for the Qwen3-1.7B student over 180,000 steps, under the settings in Appendix~\ref{app:distill}.
Both runs use the same distillation objective, data, optimizer, and training schedule.
The displayed task loss excludes the structural coupling penalty.
These results show that introducing structured sparsity preserves stable convergence across the evaluated settings, despite a modest increase in task loss relative to vanilla training.
In the Llama pretraining experiments, S-SALAAD also maintains loss trajectories comparable to those of unstructured SALAAD.

\subsection{Learned Sparsity Patterns}
\label{app:training_patterns}

\begin{figure}[htbp]
  \centering
  \includegraphics[width=\linewidth]{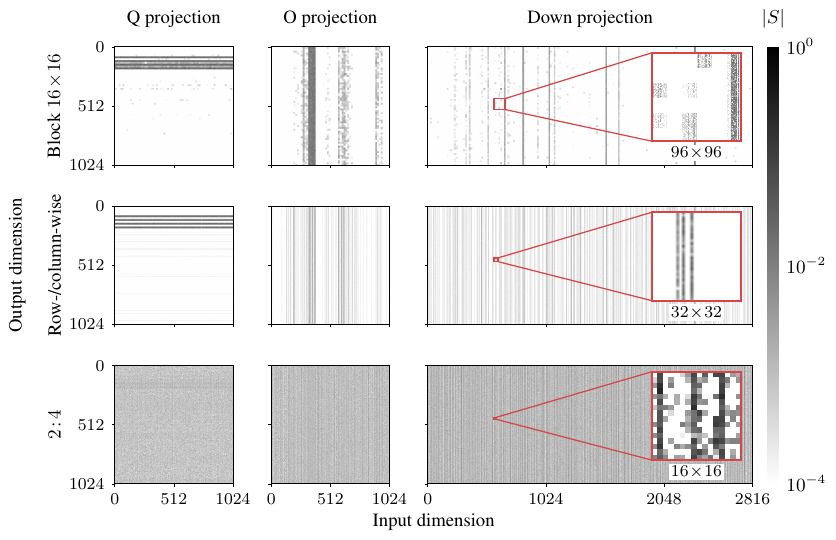}
  \caption{Learned sparse components $S$ in layer~16 of Llama-350M. Rows show block-$16\!\times\!16$, row-/column-wise, and $2\!:\!4$ training; columns show the Q, O, and down projections. Insets enlarge the red-boxed regions of each down projection on the same color scale; their labels give the crop dimensions in weights. The displayed matrices are taken directly from the trained checkpoints; the $2\!:\!4$ row shows the learned components before any additional structural projection.}
  \label{fig:training_350m_heatmaps}
\end{figure}

\begin{figure}[htbp]
  \centering
  \includegraphics[width=\linewidth]{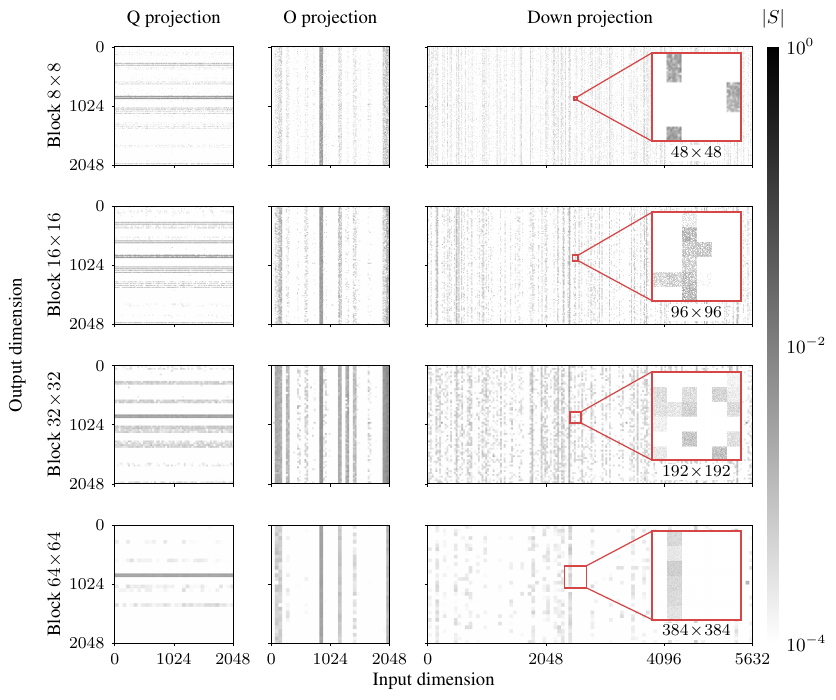}
  \caption{Learned sparse components $S$ in layer~16 of Llama-1B across four block sizes. Columns show the Q, O, and down projections. Insets enlarge the red-boxed down-projection regions on the same color scale; labels give the crop dimensions in weights. Each inset contains $6\!\times\!6$ native blocks.}
  \label{fig:training_1b_heatmaps}
\end{figure}

\begin{figure}[htbp]
  \centering
  \includegraphics[width=\linewidth]{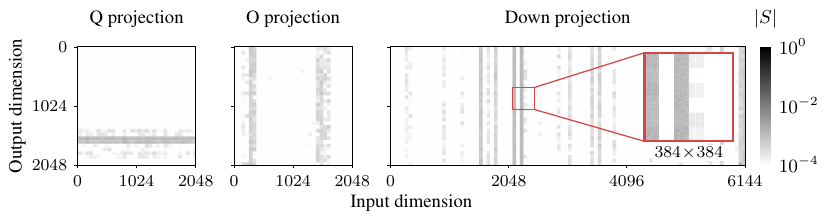}
  \caption{Learned sparse components $S$ in layer~16 of Qwen3-1.7B trained with block-$64\!\times\!64$ S-SALAAD. Columns show the Q, O, and down projections. The down-projection inset enlarges the red-boxed $384\!\times\!384$ region on the same color scale, showing $6\!\times\!6$ native blocks.}
  \label{fig:training_qwen_heatmaps}
\end{figure}

All panels in Figures~\ref{fig:training_350m_heatmaps}, \ref{fig:training_1b_heatmaps}, and~\ref{fig:training_qwen_heatmaps} preserve matrix aspect ratios and share a logarithmic scale for element-wise $|S|$, clipped to $[10^{-4},1]$, with zeros shown in white.

Figure~\ref{fig:training_350m_heatmaps} displays the learned sparse components of randomly selected projections in Llama-350M for the three training variants in Figure~\ref{fig:training_350m_loss}.
For the $2\!:\!4$ training variant, the displayed components have approximately 50\% density, while occupancy varies across individual four-entry groups.
The regularizer and I-Controller encourage the $N\!:\!M$ target during training, while projection in EGAR Stage~II enforces at most $N$ nonzero entries per group of $M$.
Figure~\ref{fig:training_1b_heatmaps} displays the learned sparse components of randomly selected projections in Llama-1B for the four block granularities in Figure~\ref{fig:training_1b_loss}.
The insets show $6\!\times\!6$ native blocks around a common matrix location, making the local block structure visible at each granularity.
Figure~\ref{fig:training_qwen_heatmaps} shows that S-SALAAD induces the intended $64\!\times\!64$ block structure in randomly selected projections of Qwen3-1.7B during distillation.

The block-sparse components in Figure~\ref{fig:training_350m_heatmaps} also exhibit row- or column-aligned clustering resembling that of the row-/column-wise variant.
This organization persists across block granularities (Figure~\ref{fig:training_1b_heatmaps}) and is also observed in Qwen3-1.7B under a different model architecture and training setting (Figure~\ref{fig:training_qwen_heatmaps}).
We conjecture that this recurring organization reflects an intrinsic sparsity structure associated with semantic representations in Transformers.
A detailed investigation of this hypothesis is beyond the scope of this work.

\FloatBarrier
\subsection{Paired Sparsity Patterns}
\label{app:paired_sparsity}

\begin{figure}[!htbp]
  \centering
  \includegraphics[width=\linewidth]{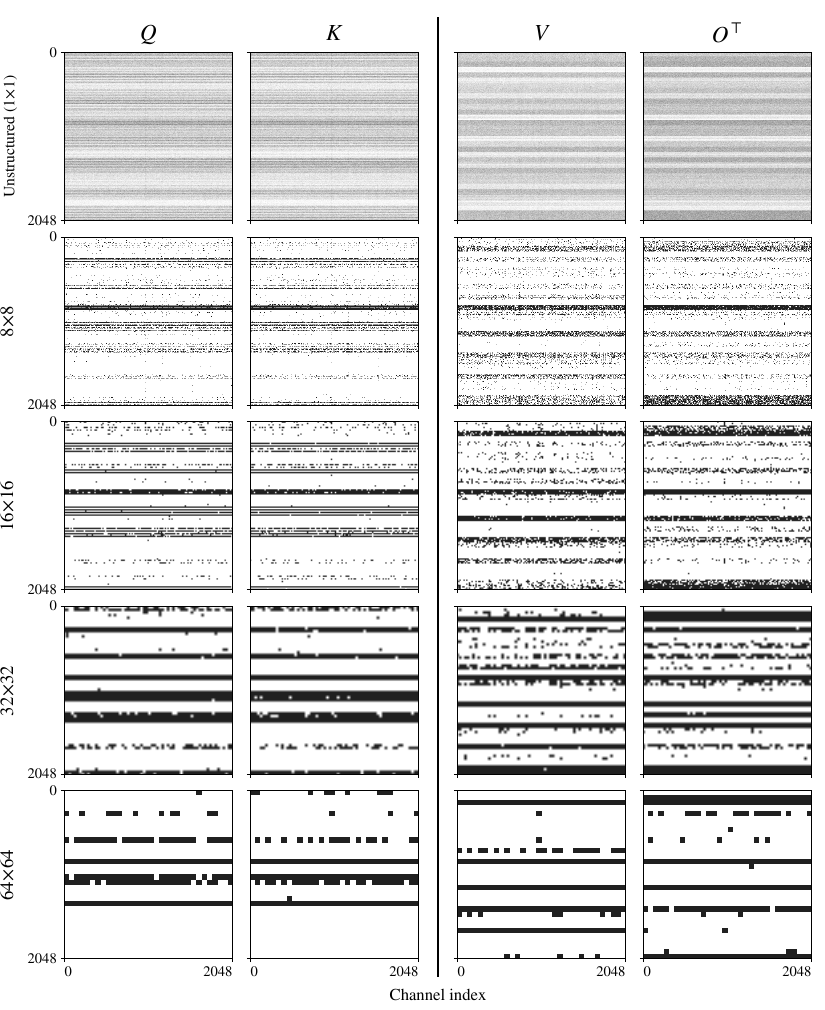}
  \caption{Paired sparsity patterns in a randomly selected layer of Llama-1B. Rows show unstructured sparsity ($1\!\times\!1$) followed by four block sizes. Dark cells are nonzero and white cells are exactly zero. The O component is transposed so that its head slices align with those of V along rows.}
  \label{fig:paired_sparsity}
\end{figure}

Beyond the structure within individual projections, the learned sparse components exhibit paired patterns between Q and K and between V and O within the same layer, as illustrated in Figure~\ref{fig:paired_sparsity}.
The figure shows the sparse structure of a randomly selected layer of Llama-1B, with its five rows corresponding to unstructured sparsity ($1\!\times\!1$) and four block sizes.

\textbf{Paired sparsity within a layer.}
In each row of Figure~\ref{fig:paired_sparsity}, nonzero entries in Q and K concentrate at similar row positions, while the nonzero regions in V and $O^{\top}$ also show correspondence across matching head slices.
Each layer has 32 attention heads of dimension 64.
Q/K pairing compares the output-row slices belonging to the same head, while V/O pairing compares a V output-row slice with the corresponding O input-column slice.
This consistency concerns the distribution of nonzero entries over heads, it does not require identical block supports within paired slices.

To examine whether this pattern extends to other layers, we analyze all 24 layers of the five checkpoints.
For each attention projection $\mathrm{P}\!\in\!\{\mathrm{Q},\mathrm{K},\mathrm{V},\mathrm{O}\}$, let $d_{\mathrm{P},h}$ denote the fraction of nonzero entries in the slice of its sparse component corresponding to head $h$.
For each pair $(\mathrm{A}_1,\mathrm{A}_2)\!\in\!\{(\mathrm{Q},\mathrm{K}),(\mathrm{V},\mathrm{O})\}$, the Pearson correlation within a layer is
\[
  r_{\mathrm{A}_1,\mathrm{A}_2} =
  \frac{\sum_{h=1}^{H}(d_{\mathrm{A}_1,h}-\bar d_{\mathrm{A}_1})(d_{\mathrm{A}_2,h}-\bar d_{\mathrm{A}_2})}
       {\sqrt{\sum_{h=1}^{H}(d_{\mathrm{A}_1,h}-\bar d_{\mathrm{A}_1})^2}\,
        \sqrt{\sum_{h=1}^{H}(d_{\mathrm{A}_2,h}-\bar d_{\mathrm{A}_2})^2}},
\]
where $H\!=\!32$ and $\bar d_{\mathrm{P}}\!=\!H^{-1}\sum_{h=1}^{H}d_{\mathrm{P},h}$.
Layers in which either projection has an entirely zero sparse component are excluded from the statistics for that pair, since the Pearson correlation is undefined.
Table~\ref{tab:paired_sparsity} reports the correlations across valid layers as mean $\pm$ standard deviation, showing strong average Q/K and V/O correlations at every block size.
\begin{table}[h]
  \centering
  \caption{Mean $\pm$ sample standard deviation of layer-wise Pearson correlations of head nonzero fractions.  Parentheses give the number $n$ of valid layers for each pair, excluding layers in which either projection has an entirely zero sparse component.}
  \label{tab:paired_sparsity}
  \begin{tabular}{lcc}
    \toprule
    Block size & Q/K correlation (layers) & V/O correlation (layers)\\
    \midrule
    Unstructured ($1\!\times\!1$) & $0.971\pm0.096$ (24) & $0.976\pm0.024$ (24)\\
    $8\!\times\!8$   & $0.986\pm0.025$ (23) & $0.962\pm0.051$ (24)\\
    $16\!\times\!16$ & $0.979\pm0.039$ (23) & $0.948\pm0.076$ (24)\\
    $32\!\times\!32$ & $0.980\pm0.018$ (21) & $0.937\pm0.078$ (24)\\
    $64\!\times\!64$ & $0.967\pm0.022$ (21) & $0.909\pm0.127$ (24)\\
    \bottomrule
  \end{tabular}
\end{table}
These results indicate that the sparse components of Q/K and V/O tend to concentrate on corresponding heads within the same layer.

\textbf{Consistency of dominant heads across granularities.}
The rows of Figure~\ref{fig:paired_sparsity} show that the main concentrations of nonzero entries occupy similar head positions across sparsity granularities.
To quantify this consistency, we compare the same projection $\mathrm{P}\!\in\!\{\mathrm{Q},\mathrm{K},\mathrm{V},\mathrm{O}\}$ across granularities.
Let $d_{\mathrm{P},h}^{(b)}$ denote the nonzero fraction of head $h$ at granularity $b\!\times\!b$, where $b\!\in\!\{1,8,16,32,64\}$.
At each granularity, we select the four heads with the largest nonzero fractions and denote their index set by $T_{\mathrm{P}}^{(b)}$.
For two granularities $b$ and $b'$, the number of shared heads is $|T_{\mathrm{P}}^{(b)}\cap T_{\mathrm{P}}^{(b')}|$.
We include a layer comparison only when both granularities have at least four nonzero head slices and no tie between the fourth and fifth positions.

Table~\ref{tab:projection_top4_overlap} reports the mean and sample standard deviation across valid layers separately for each of the ten granularity pairs and each projection.
\begin{table}[h]
  \centering
  \caption{Number of shared head indices among the four heads with the largest nonzero fractions, comparing the same projection across each pair of sparsity granularities. Parentheses give the number of valid layers.}
  \label{tab:projection_top4_overlap}
  \small
  \setlength{\tabcolsep}{4pt}
  \begin{tabular}{lcccc}
    \toprule
    Granularity pair & Q & K & V & O\\
    \midrule
    $1\!\times\!1$ / $8\!\times\!8$ & $3.04\pm0.77$ (23) & $2.73\pm0.88$ (22) & $3.08\pm0.72$ (24) & $2.88\pm0.61$ (24)\\
    $1\!\times\!1$ / $16\!\times\!16$ & $3.11\pm0.88$ (19) & $2.72\pm0.83$ (18) & $2.92\pm0.78$ (24) & $2.92\pm0.72$ (24)\\
    $1\!\times\!1$ / $32\!\times\!32$ & $3.29\pm0.77$ (17) & $3.00\pm0.61$ (17) & $3.00\pm0.63$ (16) & $2.91\pm0.83$ (11)\\
    $1\!\times\!1$ / $64\!\times\!64$ & $3.29\pm0.73$ (14) & $2.75\pm0.68$ (16) & $3.12\pm0.81$ (16) & $2.93\pm0.73$ (14)\\
    $8\!\times\!8$ / $16\!\times\!16$ & $3.37\pm0.60$ (19) & $3.06\pm0.73$ (18) & $3.25\pm0.61$ (24) & $3.25\pm0.68$ (24)\\
    $8\!\times\!8$ / $32\!\times\!32$ & $3.12\pm0.60$ (17) & $2.94\pm0.66$ (17) & $3.38\pm0.50$ (16) & $3.27\pm0.65$ (11)\\
    $8\!\times\!8$ / $64\!\times\!64$ & $3.36\pm0.63$ (14) & $3.06\pm0.68$ (16) & $3.38\pm0.62$ (16) & $3.29\pm0.73$ (14)\\
    $16\!\times\!16$ / $32\!\times\!32$ & $3.44\pm0.81$ (16) & $3.33\pm0.90$ (15) & $3.38\pm0.50$ (16) & $3.36\pm0.50$ (11)\\
    $16\!\times\!16$ / $64\!\times\!64$ & $3.58\pm0.67$ (12) & $3.15\pm0.69$ (13) & $3.25\pm0.58$ (16) & $3.50\pm0.65$ (14)\\
    $32\!\times\!32$ / $64\!\times\!64$ & $3.31\pm0.75$ (13) & $3.20\pm0.68$ (15) & $3.31\pm0.63$ (13) & $3.22\pm0.44$ (9)\\
    \bottomrule
  \end{tabular}
\end{table}

This consistency concerns the overlap of dominant heads and does not imply identical sets of nonzero heads or identical connections within them.
Among the eligible layers, the dominant head locations within each projection are broadly consistent across the examined sparsity granularities.

\subsection{Effective-Rank and Sparsity Ratios}
\label{app:training_rank_sparsity}

\begin{figure}[htbp]
  \centering
  \includegraphics[width=\linewidth]{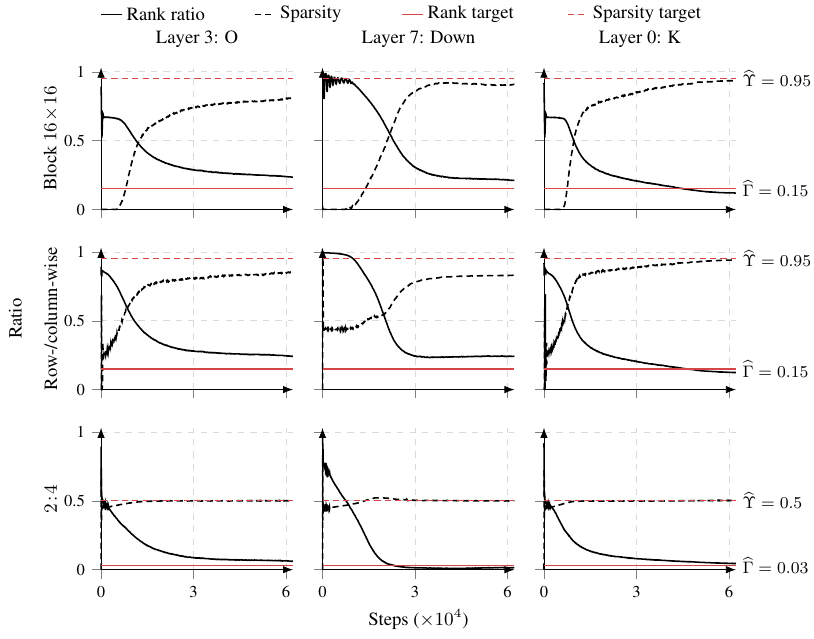}
  \caption{Effective-rank ratio and sparsity during Llama-350M pretraining. Rows correspond to block-$16\!\times\!16$, row-/column-wise, and $2\!:\!4$ sparsity; columns show the O projection in layer~3, down projection in layer~7, and K projection in layer~0. Black solid and dashed curves show the rank ratio and sparsity (the zero fraction), respectively; red lines show the corresponding controller targets. The statistics are unsmoothed.}
  \label{fig:training_350m_rank_sparsity}
\end{figure}

\begin{figure}[htbp]
  \centering
  \includegraphics[width=\linewidth]{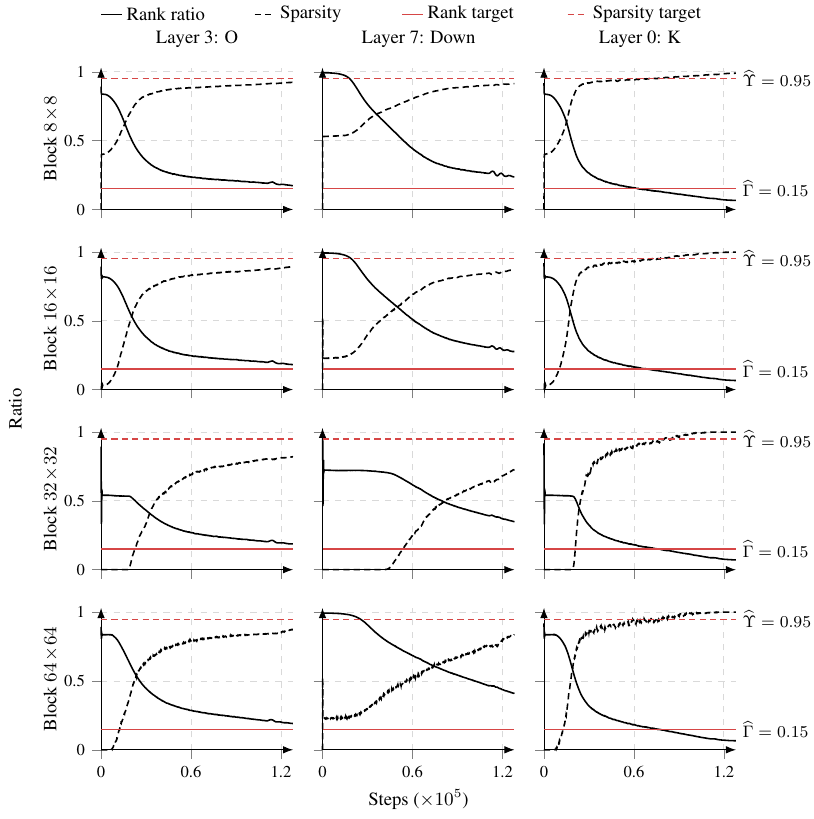}
  \caption{Effective-rank ratio and sparsity during Llama-1B pretraining. Rows correspond to block sizes $8$, $16$, $32$, and $64$; columns show the O projection in layer~3, down projection in layer~7, and K projection in layer~0. Black solid and dashed curves show the rank ratio and sparsity, respectively; red lines indicate their targets. The statistics are unsmoothed.}
  \label{fig:training_1b_rank_sparsity}
\end{figure}

\begin{figure}[htbp]
  \centering
  \includegraphics[width=\linewidth]{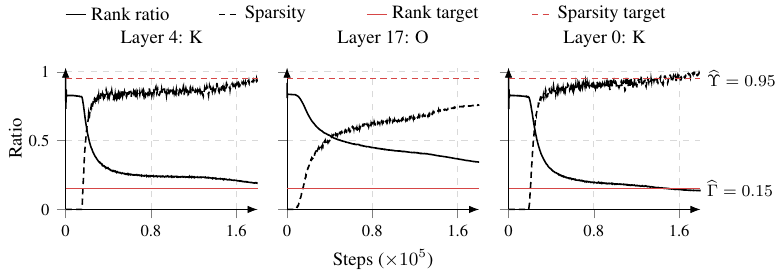}
  \caption{Effective-rank ratio and sparsity during Qwen3-1.7B distillation with block-$64\!\times\!64$ S-SALAAD. Columns show the randomly selected K projection in layer~4, O projection in layer~17, and K projection in layer~0. Black solid and dashed curves show the rank ratio and sparsity, respectively; red lines indicate their targets. The statistics are unsmoothed.}
  \label{fig:training_qwen_rank_sparsity}
\end{figure}

Figures~\ref{fig:training_350m_rank_sparsity}, \ref{fig:training_1b_rank_sparsity}, and~\ref{fig:training_qwen_rank_sparsity} show how the effective-rank ratio and sparsity of sampled projections progressively approach their prescribed targets during training.
We use a sparsity target of $\widehat{\Upsilon}\!=\!0.95$ and an effective-rank ratio target of $\widehat{\Gamma}\!=\!0.15$.
For the $2\!:\!4$ pattern, the fixed $50\%$ sparsity level gives $\widehat{\Upsilon}\!=\!0.5$, and we correspondingly lower the rank target to $\widehat{\Gamma}\!=\!0.03$.

Figure~\ref{fig:training_350m_rank_sparsity} compares block-$16\!\times\!16$, row-/column-wise, and $2\!:\!4$ sparsity on Llama-350M.
Figure~\ref{fig:training_1b_rank_sparsity} compares four block granularities on Llama-1B.
Figure~\ref{fig:training_qwen_rank_sparsity} shows the corresponding trajectories for three randomly selected projections during Qwen3-1.7B distillation with block-$64\!\times\!64$ S-SALAAD.


Across the evaluated sparsity patterns, model sizes, and architectures, the weight components progressively approach the prescribed structure, demonstrating the effectiveness of S-SALAAD's structure induction.
The trajectories also reveal distinct convergence behavior across layers and projection types.

\section{Additional Results}
\label{app:additional_deployment}

\subsection{Pareto Frontiers on Other Hardware}
\label{app:additional_hardware}
\begin{figure}[htbp]
  \centering
  \begin{subfigure}[b]{0.49\linewidth}
    \centering
    \includegraphics[width=\linewidth]{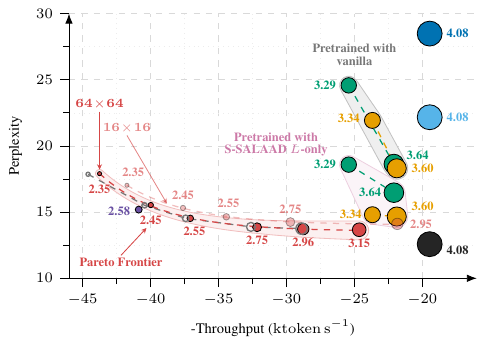}
    \caption{Llama-1B.}
    \label{fig:e2e_2080ti_1b}
  \end{subfigure}\hfill
  \begin{subfigure}[b]{0.49\linewidth}
    \centering
    \includegraphics[width=\linewidth]{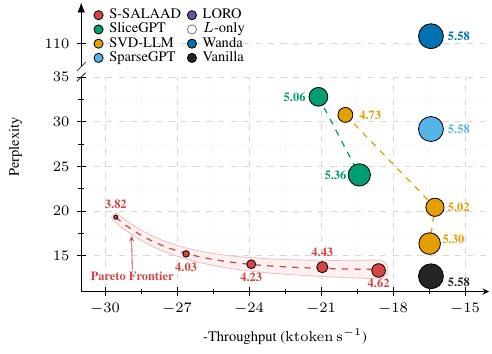}
    \caption{Qwen3-1.7B.}
    \label{fig:e2e_2080ti_qwen}
  \end{subfigure}
  \caption{Perplexity against throughput on an RTX~2080~Ti; marker radius linearly encodes peak GPU memory usage, with values annotated in GB. Lighter red denotes the $16\!\times\!16$ variant. Both panels share the legend in the right panel.}
  \label{fig:e2e_2080ti}
\end{figure}
\begin{figure}[h]
  \centering
  \begin{subfigure}[b]{0.49\linewidth}
    \centering
    \includegraphics[width=\linewidth]{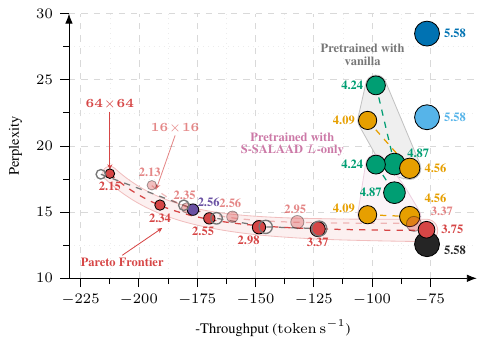}
    \caption{Llama-1B.}
    \label{fig:e2e_cpu_1b}
  \end{subfigure}\hfill
  \begin{subfigure}[b]{0.49\linewidth}
    \centering
    \includegraphics[width=\linewidth]{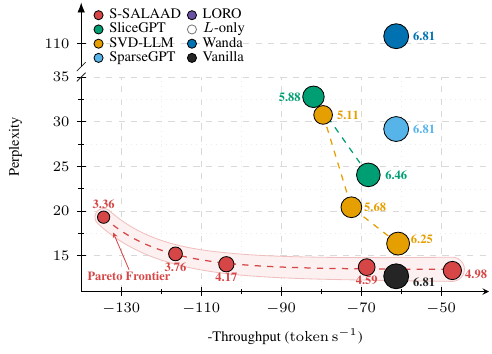}
    \caption{Qwen3-1.7B.}
    \label{fig:e2e_cpu_qwen}
  \end{subfigure}
  \caption{Perplexity against throughput on 4$\times$ AMD~EPYC~7662 CPUs (batch size~8, sequence length~64); marker radius is proportional to peak RAM usage, with values annotated in GB. Lighter red denotes the $16\!\times\!16$ variant. Both panels share the legend in the right panel.}
  \label{fig:e2e_cpu}
\end{figure}
Figures~\ref{fig:e2e_2080ti} and~\ref{fig:e2e_cpu} extend the deployment evaluation to an RTX~2080~Ti and 4$\times$ AMD~EPYC~7662 CPUs.
Relative to vanilla inference, block-$64\!\times\!64$ S-SALAAD at a 500M budget reduces peak GPU memory usage by 42.4\% and achieves $2.25\times$ throughput on Llama-1B with the RTX~2080~Ti (Figure~\ref{fig:e2e_2080ti_1b}).
For Qwen3-1.7B at 800M, the corresponding gains are 31.5\% lower peak GPU memory usage and $1.80\times$ throughput (Figure~\ref{fig:e2e_2080ti_qwen}).
On CPU, these compressed variants reduce peak RAM usage by 61.5\% with $2.78\times$ throughput for Llama-1B (Figure~\ref{fig:e2e_cpu_1b}), and by 50.6\% with $2.20\times$ throughput for Qwen3-1.7B (Figure~\ref{fig:e2e_cpu_qwen}).
These results demonstrate that S-SALAAD's memory and inference gains extend across the evaluated hardware platforms.
Detailed numerical results are provided in Appendix~\ref{app:pareto_numerical_results}.

The deployment conclusions of Sections~\ref{sec:results} and~\ref{sec:ablation_studies} also extend to these settings.

\textbf{Synergy with post-training compression.}
Applying SliceGPT and SVD-LLM to $L$-only checkpoints pretrained with S-SALAAD yields lower perplexity at the same inference cost than applying them to vanilla-pretrained checkpoints.
This benefit persists on both GPU and CPU (Figures~\ref{fig:e2e_2080ti_1b} and~\ref{fig:e2e_cpu_1b}), demonstrating that the learned compressibility remains useful across hardware.

\textbf{Block granularity and deployment trade-offs.}
On Llama-1B, the $16\!\times\!16$ variant achieves lower perplexity than $L$-only under tight compression budgets (Figure~\ref{fig:pvp_1b}).
Its additional inference overhead, however, limits the deployment gains on both GPU and CPU (Figures~\ref{fig:e2e_2080ti_1b} and~\ref{fig:e2e_cpu_1b}).
Increasing the block size reduces this overhead, allowing $64\!\times\!64$ to closely track the $L$-only perplexity--memory--latency trade-offs on the RTX~2080~Ti and improve on them at CPU operating points.
These trends support the conclusion that retaining $S$ can improve deployment trade-offs beyond $L$-only when block granularity is matched to the target hardware, as further illustrated in Appendix~\ref{app:block_granularity_pareto}.

\subsection{Qwen Inference Under a Fixed Memory Cap}
\label{app:memory_cap}

\begin{figure}[htbp]
  \centering
  \includegraphics[width=0.60\linewidth]{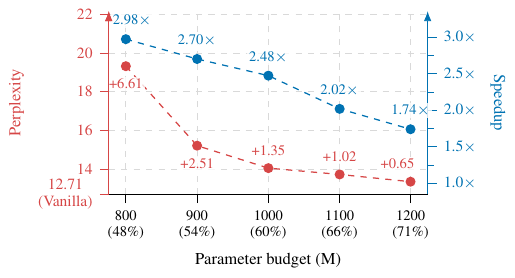}
  \caption{Compressed Qwen3-1.7B: speedup and perplexity under a \qty{3.4}{\giga\byte} peak-memory cap. We increase the batch size until the cap.}
  \label{fig:qwen_distill}
  \label{fig:qwen_distill_gpu}
\end{figure}

Under a constrained GPU memory budget, S-SALAAD's memory savings also enable larger inference batches, providing additional throughput gains.
Figure~\ref{fig:qwen_distill_gpu} evaluates Qwen3-1.7B on the Quadro~RTX~6000 under a fixed \qty{3.4}{\giga\byte} peak-memory cap, with the batch size increased for each model within this limit.
At a 1B parameter budget, S-SALAAD achieves $2.48\times$ the throughput of dense inference, with perplexity increasing from 12.71 to 14.06.
These results show how elastic compression allows limited GPU memory to support higher throughput while offering a choice of model quality.

\section{Block Granularity and the Pareto Frontier}
\label{app:block_granularity_pareto}
\begin{figure}[H]
  \centering
  \captionsetup[subfigure]{skip=2pt,justification=centering}
  \newlength{\blockparetoheight}
  \setlength{\blockparetoheight}{0.38\linewidth}
  \includegraphics{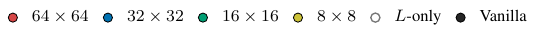}
  \par\vspace{4pt}
  \begin{subfigure}[t]{0.49\linewidth}
    \vspace{0pt}
    \centering
    \includegraphics[height=\blockparetoheight]{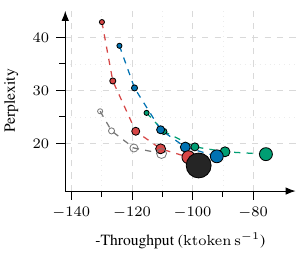}
    \caption{Llama-350M, Quadro RTX 6000.}
    \label{fig:block_speed_350m}
  \end{subfigure}\hfill
  \begin{subfigure}[t]{0.49\linewidth}
    \vspace{0pt}
    \centering
    \includegraphics[height=\blockparetoheight]{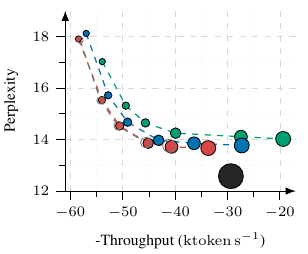}
    \caption{Llama-1B, Quadro RTX 6000.}
    \label{fig:block_speed_1b}
  \end{subfigure}
  \par\vspace{5pt}
  \begin{subfigure}[t]{0.49\linewidth}
    \vspace{0pt}
    \centering
    \includegraphics[height=\blockparetoheight]{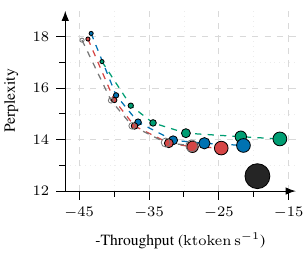}
    \caption{Llama-1B, RTX 2080 Ti.}
    \label{fig:block_speed_1b_2080ti}
  \end{subfigure}\hfill
  \begin{subfigure}[t]{0.49\linewidth}
    \vspace{0pt}
    \centering
    \includegraphics[height=\blockparetoheight]{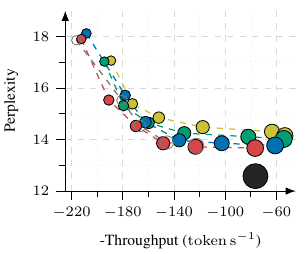}
    \caption{Llama-1B, CPU.}
    \label{fig:block_speed_1b_cpu}
  \end{subfigure}
  \caption{Perplexity against throughput on Llama for different block granularities, including the \mbox{$L$-only} baseline. Marker radius encodes peak GPU memory with min--max scaling or peak RAM with proportional scaling, per panel.}
  \label{fig:block_speed}
\end{figure}

Figure~\ref{fig:block_speed} extends the analysis in Section~\ref{sec:ablation_studies} by comparing the deployment trade-offs of different block granularities and the $L$-only special case across model scales and hardware.
Finer blocks better preserve quality under tight compression budgets, but their additional inference overhead limits the deployment gains.
On Llama-350M, coarser blocks incur a pronounced quality penalty despite their higher throughput (Figure~\ref{fig:block_speed_350m}).
On Llama-1B, the coarsest blocks retain competitive quality and closely track the $L$-only deployment trade-offs on both GPUs (Figures~\ref{fig:block_speed_1b} and~\ref{fig:block_speed_1b_2080ti}).
On CPU, the same granularity also provides operating points that improve on $L$-only (Figure~\ref{fig:block_speed_1b_cpu}).
Together with Figure~\ref{fig:block_scaling}, these trends indicate that larger models tolerate coarser structure, enabling more efficient inference while preserving elastic compressibility.
They also support the main-text conclusion that retaining $S$ can improve the perplexity--memory--latency trade-offs beyond $L$-only when block granularity is matched to the model and deployment hardware.
Numerical results for Figure~\ref{fig:block_speed} are provided in Appendix~\ref{app:pareto_numerical_results}.

\section{Numerical Results}
\label{app:pareto_numerical_results}

Tables~\ref{tab:pareto_numeric_rtx6000}, \ref{tab:pareto_numeric_2080ti}, and~\ref{tab:pareto_numeric_cpu} list all operating points displayed in Figures~\ref{fig:e2e}, \ref{fig:e2e_2080ti}, and~\ref{fig:e2e_cpu}, respectively.
Table~\ref{tab:block_pareto_numeric} lists the operating points in Figure~\ref{fig:block_speed} not already reported in Tables~\ref{tab:pareto_numeric_rtx6000}, \ref{tab:pareto_numeric_2080ti}, and~\ref{tab:pareto_numeric_cpu}.



\begin{table}[htbp]
  \centering
  \small
  \setlength{\tabcolsep}{6pt}
  \renewcommand{\arraystretch}{1.05}
  \caption{Numerical results for Figure~\ref{fig:e2e} on a Quadro~RTX~6000. FP16; batch size $\times$ sequence length is $128\!\times\!256$ for Llama and $32\!\times\!512$ for Qwen.
  Arrows indicate preferred directions. Within each model, \textbf{bold} and \underline{underlined} values mark the best and second-best distinct values per metric across the listed settings, at the displayed precision.
  Blue rows denote S-SALAAD, including its $L$-only special case.
  For the Llama baselines, (V) and (L) denote compression of a vanilla-pretrained and of an S-SALAAD $L$-only checkpoint, respectively.}
  \label{tab:pareto_numeric_rtx6000}
\definecolor{paretoRowBlue}{HTML}{EAF3FF}
\renewcommand{\tabularxcolumn}[1]{m{#1}}
\begin{tabularx}{\linewidth}{>{\raggedright\arraybackslash}Xcccccc}
\toprule
\textbf{Method} & \multicolumn{1}{c}{\textbf{Budget}} & \multicolumn{1}{c}{\textbf{Params.}$\downarrow$} & \multicolumn{1}{c}{\textbf{Perplexity}$\downarrow$} & \multicolumn{1}{c}{\textbf{Peak mem.}$\downarrow$} & \multicolumn{1}{c}{\textbf{Latency}$\downarrow$} & \multicolumn{1}{c}{\textbf{Throughput}$\uparrow$} \\
 & \multicolumn{1}{c}{(M)} & \multicolumn{1}{c}{(M)} & \multicolumn{1}{c}{} & \multicolumn{1}{c}{(GB)} & \multicolumn{1}{c}{(ms)} & \multicolumn{1}{c}{(\si{\kilo\token\per\second})} \\
\midrule
\multicolumn{7}{c}{\textbf{Llama-1B}} \\
\midrule
Vanilla & -- & 1364.8 & \textbf{12.58} & 4.08 & 1116 & 29.35 \\ 
\addlinespace[2pt]
\rowcolor{paretoRowBlue}
 & 500 & \textbf{500.0} & 17.91 & \textbf{2.35} & \textbf{560} & \textbf{58.47} \\ 
\rowcolor{paretoRowBlue}
 & 550 & \underline{550.0} & 15.54 & \underline{2.45} & 607 & 53.99 \\ 
\rowcolor{paretoRowBlue}
 & 600 & 600.0 & 14.53 & 2.55 & 647 & 50.61 \\ 
\rowcolor{paretoRowBlue}
 & 700 & 700.0 & 13.86 & 2.75 & 726 & 45.16 \\ 
\rowcolor{paretoRowBlue}
 & 800 & 800.0 & 13.72 & 2.96 & 805 & 40.70 \\ 
\rowcolor{paretoRowBlue}
\multirow{-6}{=}{\shortstack[l]{S-SALAAD\\($64\!\times\!64$)}} & 900 & 895.0 & \underline{13.67} & 3.15 & 973 & 33.68 \\ 
\addlinespace[2pt]
\rowcolor{paretoRowBlue}
 & 500 & \textbf{500.0} & 17.03 & \textbf{2.35} & 608 & 53.93 \\ 
\rowcolor{paretoRowBlue}
 & 550 & \underline{550.0} & 15.32 & \underline{2.45} & 663 & 49.43 \\ 
\rowcolor{paretoRowBlue}
 & 600 & 600.0 & 14.65 & 2.55 & 717 & 45.69 \\ 
\rowcolor{paretoRowBlue}
 & 700 & 700.0 & 14.25 & 2.75 & 821 & 39.91 \\ 
\rowcolor{paretoRowBlue}
\multirow{-5}{=}{\shortstack[l]{S-SALAAD\\($16\!\times\!16$)}} & 800 & 800.0 & 14.11 & 2.95 & 1194 & 27.44 \\ 
\addlinespace[2pt]
\rowcolor{paretoRowBlue}
 & 500 & \textbf{500.0} & 17.86 & \textbf{2.35} & \underline{563} & \underline{58.17} \\ 
\rowcolor{paretoRowBlue}
 & 550 & \underline{550.0} & 15.52 & \underline{2.45} & 604 & 54.25 \\ 
\rowcolor{paretoRowBlue}
 & 600 & 600.0 & 14.53 & 2.55 & 644 & 50.90 \\ 
\rowcolor{paretoRowBlue}
 & 700 & 700.0 & 13.89 & 2.75 & 718 & 45.61 \\ 
\rowcolor{paretoRowBlue}
\multirow{-5}{=}{\shortstack[l]{S-SALAAD\\($L$-only)}} & 800 & 799.9 & 13.76 & 2.96 & 796 & 41.15 \\ 
\addlinespace[2pt]
LORO & -- & 616.1 & 15.19 & 2.58 & 602 & 54.45 \\ 
\addlinespace[2pt]
 & -- & 1033.4 & 24.58 & 3.29 & 888 & 36.89 \\ 
\multirow{-2}{=}{SliceGPT (V)} & -- & 1189.1 & 18.64 & 3.64 & 1021 & 32.11 \\ 
\addlinespace[2pt]
 & -- & 1033.4 & 18.59 & 3.29 & 888 & 36.89 \\ 
\multirow{-2}{=}{SliceGPT (L)} & -- & 1189.1 & 16.47 & 3.64 & 1021 & 32.11 \\ 
\addlinespace[2pt]
 & -- & 994.9 & 21.93 & 3.34 & 965 & 33.95 \\ 
\multirow{-2}{=}{SVD-LLM (V)} & -- & 1118.2 & 18.31 & 3.60 & 1055 & 31.07 \\ 
\addlinespace[2pt]
 & -- & 994.9 & 14.81 & 3.34 & 965 & 33.95 \\ 
\multirow{-2}{=}{SVD-LLM (L)} & -- & 1118.2 & 14.67 & 3.60 & 1055 & 31.07 \\ 
\addlinespace[2pt]
Wanda & -- & 748.3 & 28.49 & 4.08 & 1116 & 29.35 \\ 
\addlinespace[2pt]
SparseGPT & -- & 748.3 & 22.17 & 4.08 & 1116 & 29.35 \\ 
\midrule
\multicolumn{7}{c}{\textbf{Qwen3-1.7B}} \\
\midrule
Vanilla & -- & 1677.8 & \textbf{12.71} & 5.58 & 677 & 24.19 \\ 
\addlinespace[2pt]
\rowcolor{paretoRowBlue}
 & 800 & \textbf{800.0} & 19.32 & \textbf{3.82} & \textbf{418} & \textbf{39.22} \\ 
\rowcolor{paretoRowBlue}
 & 900 & \underline{900.0} & 15.21 & \underline{4.03} & \underline{455} & \underline{36.03} \\ 
\rowcolor{paretoRowBlue}
 & 1000 & 1000.0 & 14.06 & 4.23 & 493 & 33.23 \\ 
\rowcolor{paretoRowBlue}
 & 1100 & 1100.0 & 13.73 & 4.43 & 578 & 28.35 \\ 
\rowcolor{paretoRowBlue}
\multirow{-5}{=}{\shortstack[l]{S-SALAAD\\($64\!\times\!64$)}} & 1200 & 1199.5 & \underline{13.36} & 4.62 & 671 & 24.42 \\ 
\addlinespace[2pt]
 & -- & 1449.1 & 32.79 & 5.06 & 544 & 30.10 \\ 
\multirow{-2}{=}{SliceGPT} & -- & 1591.8 & 24.05 & 5.36 & 590 & 27.75 \\ 
\addlinespace[2pt]
 & -- & 1254.6 & 30.76 & 4.73 & 585 & 28.02 \\ 
 & -- & 1395.4 & 20.43 & 5.02 & 653 & 25.09 \\ 
\multirow{-3}{=}{SVD-LLM} & -- & 1536.4 & 16.37 & 5.30 & 701 & 23.38 \\ 
\addlinespace[2pt]
Wanda & -- & 973.2 & 113.29 & 5.58 & 677 & 24.19 \\ 
\addlinespace[2pt]
SparseGPT & -- & 973.2 & 29.18 & 5.58 & 677 & 24.19 \\ 
\bottomrule
\end{tabularx}

\end{table}

\begin{table}[htbp]
  \centering
  \small
  \setlength{\tabcolsep}{6pt}
  \renewcommand{\arraystretch}{1.05}
  \caption{Numerical results for Figure~\ref{fig:e2e_2080ti} on an RTX~2080~Ti. FP16; batch size $\times$ sequence length is $128\!\times\!256$ for Llama and $32\!\times\!512$ for Qwen.
  Arrows indicate preferred directions. Within each model, \textbf{bold} and \underline{underlined} values mark the best and second-best distinct values per metric across the listed settings, at the displayed precision.
  Blue rows denote S-SALAAD, including its $L$-only special case.
  For the Llama baselines, (V) and (L) denote compression of a vanilla-pretrained and of an S-SALAAD $L$-only checkpoint, respectively.}
  \label{tab:pareto_numeric_2080ti}
\definecolor{paretoRowBlue}{HTML}{EAF3FF}
\renewcommand{\tabularxcolumn}[1]{m{#1}}
\begin{tabularx}{\linewidth}{>{\raggedright\arraybackslash}Xcccccc}
\toprule
\textbf{Method} & \multicolumn{1}{c}{\textbf{Budget}} & \multicolumn{1}{c}{\textbf{Params.}$\downarrow$} & \multicolumn{1}{c}{\textbf{Perplexity}$\downarrow$} & \multicolumn{1}{c}{\textbf{Peak mem.}$\downarrow$} & \multicolumn{1}{c}{\textbf{Latency}$\downarrow$} & \multicolumn{1}{c}{\textbf{Throughput}$\uparrow$} \\
 & \multicolumn{1}{c}{(M)} & \multicolumn{1}{c}{(M)} & \multicolumn{1}{c}{} & \multicolumn{1}{c}{(GB)} & \multicolumn{1}{c}{(ms)} & \multicolumn{1}{c}{(\si{\kilo\token\per\second})} \\
\midrule
\multicolumn{7}{c}{\textbf{Llama-1B}} \\
\midrule
Vanilla & -- & 1364.8 & \textbf{12.58} & 4.08 & 1683 & 19.47 \\ 
\addlinespace[2pt]
\rowcolor{paretoRowBlue}
 & 500 & \textbf{500.0} & 17.91 & \textbf{2.35} & \underline{749} & \underline{43.75} \\ 
\rowcolor{paretoRowBlue}
 & 550 & \underline{550.0} & 15.54 & \underline{2.45} & 820 & 39.98 \\ 
\rowcolor{paretoRowBlue}
 & 600 & 600.0 & 14.53 & 2.55 & 884 & 37.08 \\ 
\rowcolor{paretoRowBlue}
 & 700 & 700.0 & 13.86 & 2.75 & 1019 & 32.16 \\ 
\rowcolor{paretoRowBlue}
 & 800 & 800.0 & 13.72 & 2.96 & 1139 & 28.77 \\ 
\rowcolor{paretoRowBlue}
\multirow{-6}{=}{\shortstack[l]{S-SALAAD\\($64\!\times\!64$)}} & 900 & 895.0 & \underline{13.67} & 3.15 & 1329 & 24.65 \\ 
\addlinespace[2pt]
\rowcolor{paretoRowBlue}
 & 500 & \textbf{500.0} & 17.03 & \textbf{2.35} & 785 & 41.74 \\ 
\rowcolor{paretoRowBlue}
 & 550 & \underline{550.0} & 15.32 & \underline{2.45} & 871 & 37.61 \\ 
\rowcolor{paretoRowBlue}
 & 600 & 600.0 & 14.65 & 2.55 & 952 & 34.42 \\ 
\rowcolor{paretoRowBlue}
 & 700 & 700.0 & 14.25 & 2.75 & 1103 & 29.71 \\ 
\rowcolor{paretoRowBlue}
\multirow{-5}{=}{\shortstack[l]{S-SALAAD\\($16\!\times\!16$)}} & 800 & 800.0 & 14.11 & 2.95 & 1500 & 21.85 \\ 
\addlinespace[2pt]
\rowcolor{paretoRowBlue}
 & 500 & \textbf{500.0} & 17.86 & \textbf{2.35} & \textbf{735} & \textbf{44.60} \\ 
\rowcolor{paretoRowBlue}
 & 550 & \underline{550.0} & 15.52 & \underline{2.45} & 810 & 40.45 \\ 
\rowcolor{paretoRowBlue}
 & 600 & 600.0 & 14.53 & 2.55 & 876 & 37.42 \\ 
\rowcolor{paretoRowBlue}
 & 700 & 700.0 & 13.89 & 2.75 & 1005 & 32.61 \\ 
\rowcolor{paretoRowBlue}
\multirow{-5}{=}{\shortstack[l]{S-SALAAD\\($L$-only)}} & 800 & 799.9 & 13.76 & 2.96 & 1134 & 28.89 \\ 
\addlinespace[2pt]
LORO & -- & 616.1 & 15.19 & 2.58 & 802 & 40.87 \\ 
\addlinespace[2pt]
 & -- & 1033.4 & 24.58 & 3.29 & 1289 & 25.42 \\ 
\multirow{-2}{=}{SliceGPT (V)} & -- & 1189.1 & 18.64 & 3.64 & 1483 & 22.10 \\ 
\addlinespace[2pt]
 & -- & 1033.4 & 18.59 & 3.29 & 1289 & 25.42 \\ 
\multirow{-2}{=}{SliceGPT (L)} & -- & 1189.1 & 16.47 & 3.64 & 1483 & 22.10 \\ 
\addlinespace[2pt]
 & -- & 994.9 & 21.93 & 3.34 & 1384 & 23.68 \\ 
\multirow{-2}{=}{SVD-LLM (V)} & -- & 1118.2 & 18.31 & 3.60 & 1497 & 21.89 \\ 
\addlinespace[2pt]
 & -- & 994.9 & 14.81 & 3.34 & 1384 & 23.68 \\ 
\multirow{-2}{=}{SVD-LLM (L)} & -- & 1118.2 & 14.67 & 3.60 & 1497 & 21.89 \\ 
\addlinespace[2pt]
Wanda & -- & 748.3 & 28.49 & 4.08 & 1683 & 19.47 \\ 
\addlinespace[2pt]
SparseGPT & -- & 748.3 & 22.17 & 4.08 & 1683 & 19.47 \\ 
\midrule
\multicolumn{7}{c}{\textbf{Qwen3-1.7B}} \\
\midrule
Vanilla & -- & 1677.8 & \textbf{12.71} & 5.58 & 998 & 16.42 \\ 
\addlinespace[2pt]
\rowcolor{paretoRowBlue}
 & 800 & \textbf{800.0} & 19.32 & \textbf{3.82} & \textbf{554} & \textbf{29.56} \\ 
\rowcolor{paretoRowBlue}
 & 900 & \underline{900.0} & 15.21 & \underline{4.03} & \underline{615} & \underline{26.63} \\ 
\rowcolor{paretoRowBlue}
 & 1000 & 1000.0 & 14.06 & 4.23 & 685 & 23.91 \\ 
\rowcolor{paretoRowBlue}
 & 1100 & 1100.0 & 13.73 & 4.43 & 782 & 20.95 \\ 
\rowcolor{paretoRowBlue}
\multirow{-5}{=}{\shortstack[l]{S-SALAAD\\($64\!\times\!64$)}} & 1200 & 1199.5 & \underline{13.36} & 4.62 & 881 & 18.60 \\ 
\addlinespace[2pt]
 & -- & 1449.1 & 32.79 & 5.06 & 776 & 21.11 \\ 
\multirow{-2}{=}{SliceGPT} & -- & 1591.8 & 24.05 & 5.36 & 844 & 19.41 \\ 
\addlinespace[2pt]
 & -- & 1254.6 & 30.76 & 4.73 & 820 & 19.99 \\ 
 & -- & 1395.4 & 20.43 & 5.02 & 1008 & 16.26 \\ 
\multirow{-3}{=}{SVD-LLM} & -- & 1536.4 & 16.37 & 5.30 & 994 & 16.48 \\ 
\addlinespace[2pt]
Wanda & -- & 973.2 & 113.29 & 5.58 & 998 & 16.42 \\ 
\addlinespace[2pt]
SparseGPT & -- & 973.2 & 29.18 & 5.58 & 998 & 16.42 \\ 
\bottomrule
\end{tabularx}

\end{table}

\begin{table}[htbp]
  \centering
  \small
  \setlength{\tabcolsep}{6pt}
  \renewcommand{\arraystretch}{1.05}
  \caption{Numerical results for Figure~\ref{fig:e2e_cpu} on 4$\times$ AMD~EPYC~7662 CPUs. FP32; batch size~8 and sequence length~64 for both models.
  Arrows indicate preferred directions. Within each model, \textbf{bold} and \underline{underlined} values mark the best and second-best distinct values per metric across the listed settings, at the displayed precision.
  Blue rows denote S-SALAAD, including its $L$-only special case.
  For the Llama baselines, (V) and (L) denote compression of a vanilla-pretrained and of an S-SALAAD $L$-only checkpoint, respectively.}
  \label{tab:pareto_numeric_cpu}
\definecolor{paretoRowBlue}{HTML}{EAF3FF}
\renewcommand{\tabularxcolumn}[1]{m{#1}}
\begin{tabularx}{\linewidth}{>{\raggedright\arraybackslash}Xcccccc}
\toprule
\textbf{Method} & \multicolumn{1}{c}{\textbf{Budget}} & \multicolumn{1}{c}{\textbf{Params.}$\downarrow$} & \multicolumn{1}{c}{\textbf{Perplexity}$\downarrow$} & \multicolumn{1}{c}{\textbf{Peak mem.}$\downarrow$} & \multicolumn{1}{c}{\textbf{Latency}$\downarrow$} & \multicolumn{1}{c}{\textbf{Throughput}$\uparrow$} \\
 & \multicolumn{1}{c}{(M)} & \multicolumn{1}{c}{(M)} & \multicolumn{1}{c}{} & \multicolumn{1}{c}{(GB)} & \multicolumn{1}{c}{(ms)} & \multicolumn{1}{c}{(\si{\token\per\second})} \\
\midrule
\multicolumn{7}{c}{\textbf{Llama-1B}} \\
\midrule
Vanilla & -- & 1364.8 & \textbf{12.58} & 5.58 & 6705 & 76.36 \\ 
\addlinespace[2pt]
\rowcolor{paretoRowBlue}
 & 500 & \textbf{500.0} & 17.91 & \underline{2.15} & \underline{2410} & \underline{212.46} \\ 
\rowcolor{paretoRowBlue}
 & 550 & \underline{550.0} & 15.54 & 2.34 & 2681 & 190.96 \\ 
\rowcolor{paretoRowBlue}
 & 600 & 600.0 & 14.53 & 2.55 & 3016 & 169.77 \\ 
\rowcolor{paretoRowBlue}
 & 700 & 700.0 & 13.86 & 2.98 & 3449 & 148.47 \\ 
\rowcolor{paretoRowBlue}
 & 800 & 800.0 & 13.72 & 3.37 & 4154 & 123.26 \\ 
\rowcolor{paretoRowBlue}
\multirow{-6}{=}{\shortstack[l]{S-SALAAD\\($64\!\times\!64$)}} & 900 & 895.0 & \underline{13.67} & 3.75 & 6685 & 76.59 \\ 
\addlinespace[2pt]
\rowcolor{paretoRowBlue}
 & 500 & \textbf{500.0} & 17.03 & \textbf{2.13} & 2633 & 194.45 \\ 
\rowcolor{paretoRowBlue}
 & 550 & \underline{550.0} & 15.32 & 2.35 & 2853 & 179.48 \\ 
\rowcolor{paretoRowBlue}
 & 600 & 600.0 & 14.65 & 2.56 & 3202 & 159.88 \\ 
\rowcolor{paretoRowBlue}
 & 700 & 700.0 & 14.25 & 2.95 & 3878 & 132.02 \\ 
\rowcolor{paretoRowBlue}
\multirow{-5}{=}{\shortstack[l]{S-SALAAD\\($16\!\times\!16$)}} & 800 & 800.0 & 14.11 & 3.37 & 6245 & 81.98 \\ 
\addlinespace[2pt]
\rowcolor{paretoRowBlue}
 & 500 & \textbf{500.0} & 17.86 & \textbf{2.13} & \textbf{2367} & \textbf{216.32} \\ 
\rowcolor{paretoRowBlue}
 & 550 & \underline{550.0} & 15.52 & 2.34 & 2833 & 180.72 \\ 
\rowcolor{paretoRowBlue}
 & 600 & 600.0 & 14.53 & 2.56 & 3072 & 166.67 \\ 
\rowcolor{paretoRowBlue}
 & 700 & 700.0 & 13.89 & 2.95 & 3517 & 145.58 \\ 
\rowcolor{paretoRowBlue}
\multirow{-5}{=}{\shortstack[l]{S-SALAAD\\($L$-only)}} & 800 & 799.9 & 13.76 & 3.37 & 4176 & 122.60 \\ 
\addlinespace[2pt]
LORO & -- & 616.1 & 15.19 & 2.56 & 2896 & 176.79 \\ 
\addlinespace[2pt]
 & -- & 1033.4 & 24.58 & 4.24 & 5209 & 98.29 \\ 
\multirow{-2}{=}{SliceGPT (V)} & -- & 1189.1 & 18.64 & 4.87 & 5667 & 90.34 \\ 
\addlinespace[2pt]
 & -- & 1033.4 & 18.59 & 4.24 & 5209 & 98.29 \\ 
\multirow{-2}{=}{SliceGPT (L)} & -- & 1189.1 & 16.47 & 4.87 & 5667 & 90.34 \\ 
\addlinespace[2pt]
 & -- & 994.9 & 21.93 & 4.09 & 5026 & 101.86 \\ 
\multirow{-2}{=}{SVD-LLM (V)} & -- & 1118.2 & 18.31 & 4.56 & 6110 & 83.79 \\ 
\addlinespace[2pt]
 & -- & 994.9 & 14.81 & 4.09 & 5026 & 101.86 \\ 
\multirow{-2}{=}{SVD-LLM (L)} & -- & 1118.2 & 14.67 & 4.56 & 6110 & 83.79 \\ 
\addlinespace[2pt]
Wanda & -- & 748.3 & 28.49 & 5.58 & 6705 & 76.36 \\ 
\addlinespace[2pt]
SparseGPT & -- & 748.3 & 22.17 & 5.58 & 6705 & 76.36 \\ 
\midrule
\multicolumn{7}{c}{\textbf{Qwen3-1.7B}} \\
\midrule
Vanilla & -- & 1677.8 & \textbf{12.71} & 6.81 & 8363 & 61.22 \\ 
\addlinespace[2pt]
\rowcolor{paretoRowBlue}
 & 800 & \textbf{800.0} & 19.32 & \textbf{3.36} & \textbf{3810} & \textbf{134.40} \\ 
\rowcolor{paretoRowBlue}
 & 900 & \underline{900.0} & 15.21 & \underline{3.76} & \underline{4398} & \underline{116.42} \\ 
\rowcolor{paretoRowBlue}
 & 1000 & 1000.0 & 14.06 & 4.17 & 4937 & 103.70 \\ 
\rowcolor{paretoRowBlue}
 & 1100 & 1100.0 & 13.73 & 4.59 & 7464 & 68.60 \\ 
\rowcolor{paretoRowBlue}
\multirow{-5}{=}{\shortstack[l]{S-SALAAD\\($64\!\times\!64$)}} & 1200 & 1199.5 & \underline{13.36} & 4.98 & 10854 & 47.17 \\ 
\addlinespace[2pt]
 & -- & 1449.1 & 32.79 & 5.88 & 6250 & 81.92 \\ 
\multirow{-2}{=}{SliceGPT} & -- & 1591.8 & 24.05 & 6.46 & 7509 & 68.19 \\ 
\addlinespace[2pt]
 & -- & 1254.6 & 30.76 & 5.11 & 6441 & 79.50 \\ 
 & -- & 1395.4 & 20.43 & 5.68 & 7068 & 72.43 \\ 
\multirow{-3}{=}{SVD-LLM} & -- & 1536.4 & 16.37 & 6.25 & 8429 & 60.74 \\ 
\addlinespace[2pt]
Wanda & -- & 973.2 & 113.29 & 6.81 & 8363 & 61.22 \\ 
\addlinespace[2pt]
SparseGPT & -- & 973.2 & 29.18 & 6.81 & 8363 & 61.22 \\ 
\bottomrule
\end{tabularx}

\end{table}

\begin{table}[htbp]
  \centering
  \fontsize{9}{9.5}\selectfont
  \setlength{\tabcolsep}{6pt}
  \renewcommand{\arraystretch}{0.98}
  \caption{Additional numerical results for Figure~\ref{fig:block_speed}. Throughput units appear in the group headings. Arrows indicate preferred directions.
  \textbf{Bold} and \underline{underlined} values mark the best and second-best distinct values at the displayed precision within each complete plotted panel, including entries referenced in Tables~\ref{tab:pareto_numeric_rtx6000}--\ref{tab:pareto_numeric_cpu}.}
  \label{tab:block_pareto_numeric}
\renewcommand{\tabularxcolumn}[1]{m{#1}}
\begin{tabularx}{\linewidth}{>{\raggedright\arraybackslash}Xcccccc}
\toprule
\textbf{Method} & \multicolumn{1}{c}{\textbf{Budget}} & \multicolumn{1}{c}{\textbf{Params.}$\downarrow$} & \multicolumn{1}{c}{\textbf{Perplexity}$\downarrow$} & \multicolumn{1}{c}{\textbf{Peak mem.}$\downarrow$} & \multicolumn{1}{c}{\textbf{Latency}$\downarrow$} & \multicolumn{1}{c}{\textbf{Throughput}$\uparrow$} \\
 & \multicolumn{1}{c}{(M)} & \multicolumn{1}{c}{(M)} & \multicolumn{1}{c}{} & \multicolumn{1}{c}{(GB)} & \multicolumn{1}{c}{(ms)} & \multicolumn{1}{c}{} \\
\midrule
\multicolumn{7}{c}{\textbf{Llama-350M, Quadro~RTX~6000}\quad (\si{\kilo\token\per\second})} \\
\midrule
Vanilla & -- & 374.1 & \textbf{15.83} & 0.92 & 84 & 97.98 \\ 
\specialrule{0.3pt}{1pt}{1pt}
 & 170 & \textbf{170.0} & 42.90 & \textbf{0.52} & \textbf{63} & \underline{129.90} \\ 
 & 180 & \underline{180.0} & 31.78 & \underline{0.54} & \underline{65} & 126.32 \\ 
 & 200 & 200.0 & 22.29 & 0.58 & 69 & 118.75 \\ 
 & 220 & 220.0 & 18.98 & 0.62 & 74 & 110.53 \\ 
\multirow{-5}{=}{\shortstack[l]{S-SALAAD\\($64\!\times\!64$)}} & 250 & 250.0 & \underline{17.38} & 0.68 & 81 & 101.35 \\ 
\specialrule{0.3pt}{1pt}{1pt}
 & 170 & \textbf{170.0} & 38.42 & \textbf{0.52} & 66 & 124.13 \\ 
 & 180 & \underline{180.0} & 30.46 & \underline{0.54} & 69 & 119.15 \\ 
 & 200 & 200.0 & 22.57 & 0.58 & 74 & 110.53 \\ 
 & 220 & 220.0 & 19.30 & 0.62 & 80 & 102.39 \\ 
\multirow{-5}{=}{\shortstack[l]{S-SALAAD\\($32\!\times\!32$)}} & 250 & 250.0 & 17.56 & 0.68 & 89 & 91.99 \\ 
\specialrule{0.3pt}{1pt}{1pt}
 & 170 & \textbf{170.0} & 25.76 & \textbf{0.52} & 71 & 115.16 \\ 
 & 180 & \underline{180.0} & 22.25 & \underline{0.54} & 75 & 109.46 \\ 
 & 200 & 200.0 & 19.34 & 0.58 & 83 & 99.23 \\ 
 & 220 & 220.0 & 18.41 & 0.62 & 92 & 89.28 \\ 
\multirow{-5}{=}{\shortstack[l]{S-SALAAD\\($16\!\times\!16$)}} & 250 & 250.0 & 17.96 & 0.68 & 108 & 75.78 \\ 
\specialrule{0.3pt}{1pt}{1pt}
 & 170 & \textbf{170.0} & 26.06 & \textbf{0.52} & \textbf{63} & \textbf{130.49} \\ 
 & 180 & \underline{180.0} & 22.36 & \underline{0.54} & \underline{65} & 126.73 \\ 
 & 200 & 200.0 & 19.13 & 0.58 & 69 & 119.34 \\ 
\multirow{-4}{=}{\shortstack[l]{S-SALAAD\\($L$-only)}} & 220 & 220.0 & 18.09 & 0.62 & 74 & 110.23 \\ 
\midrule
\multicolumn{7}{c}{\textbf{Llama-1B, Quadro~RTX~6000}\quad (\si{\kilo\token\per\second})} \\
\midrule
 & 500 & \textbf{500.0} & 18.13 & \textbf{2.35} & 575 & 57.00 \\ 
 & 550 & \underline{550.0} & 15.72 & \underline{2.45} & 621 & 52.80 \\ 
 & 600 & 600.0 & 14.68 & 2.55 & 668 & 49.09 \\ 
 & 700 & 700.0 & 13.97 & 2.75 & 759 & 43.16 \\ 
 & 800 & 800.0 & 13.86 & 2.95 & 900 & 36.41 \\ 
\multirow{-6}{=}{\shortstack[l]{S-SALAAD\\($32\!\times\!32$)}} & 900 & 899.5 & 13.77 & 3.15 & 1201 & 27.28 \\ 
\specialrule{0.3pt}{1pt}{1pt}
\shortstack[l]{S-SALAAD\\($16\!\times\!16$)} & 900 & 900.0 & 14.03 & 3.15 & 1691 & 19.37 \\ 
\midrule
\multicolumn{7}{c}{\textbf{Llama-1B, RTX~2080~Ti}\quad (\si{\kilo\token\per\second})} \\
\midrule
 & 500 & \textbf{500.0} & 18.13 & \textbf{2.35} & 757 & 43.29 \\ 
 & 550 & \underline{550.0} & 15.72 & \underline{2.45} & 825 & 39.72 \\ 
 & 600 & 600.0 & 14.68 & 2.55 & 896 & 36.57 \\ 
 & 700 & 700.0 & 13.97 & 2.75 & 1039 & 31.54 \\ 
 & 800 & 800.0 & 13.86 & 2.95 & 1211 & 27.07 \\ 
\multirow{-6}{=}{\shortstack[l]{S-SALAAD\\($32\!\times\!32$)}} & 900 & 899.5 & 13.77 & 3.15 & 1526 & 21.47 \\ 
\specialrule{0.3pt}{1pt}{1pt}
\shortstack[l]{S-SALAAD\\($16\!\times\!16$)} & 900 & 900.0 & 14.03 & 3.15 & 2016 & 16.25 \\ 
\midrule
\multicolumn{7}{c}{\textbf{Llama-1B, AMD~EPYC~7662}\quad (\si{\token\per\second})} \\
\midrule
 & 500 & \textbf{500.0} & 18.13 & 2.17 & 2455 & 208.51 \\ 
 & 550 & \underline{550.0} & 15.72 & 2.35 & 2874 & 178.16 \\ 
 & 600 & 600.0 & 14.68 & 2.57 & 3154 & 162.34 \\ 
 & 700 & 700.0 & 13.97 & 2.95 & 3766 & 135.94 \\ 
 & 800 & 800.0 & 13.86 & 3.35 & 4987 & 102.67 \\ 
\multirow{-6}{=}{\shortstack[l]{S-SALAAD\\($32\!\times\!32$)}} & 900 & 899.5 & 13.77 & 3.77 & 8394 & 60.99 \\ 
\specialrule{0.3pt}{1pt}{1pt}
\shortstack[l]{S-SALAAD\\($16\!\times\!16$)} & 900 & 900.0 & 14.03 & 3.81 & 9443 & 54.22 \\ 
\specialrule{0.3pt}{1pt}{1pt}
 & 500 & \textbf{500.0} & 17.07 & \underline{2.14} & 2705 & 189.25 \\ 
 & 550 & \underline{550.0} & 15.39 & 2.37 & 2968 & 172.53 \\ 
 & 600 & 600.0 & 14.85 & 2.56 & 3371 & 151.90 \\ 
 & 700 & 700.0 & 14.48 & 2.95 & 4356 & 117.55 \\ 
 & 800 & 800.0 & 14.31 & 3.39 & 8070 & 63.44 \\ 
\multirow{-6}{=}{\shortstack[l]{S-SALAAD\\($8\!\times\!8$)}} & 900 & 866.0 & 14.15 & 3.64 & 9596 & 53.35 \\ 
\bottomrule
\end{tabularx}

\end{table}

\end{document}